\documentclass{article} %
\usepackage{iclr2027_conference,times}

\usepackage{amsmath,amsfonts,bm}

\def\eqref#1{equation~\ref{#1}}

\def\1{\bm{1}}

\DeclareMathAlphabet{\mathsfit}{\encodingdefault}{\sfdefault}{m}{sl}
\SetMathAlphabet{\mathsfit}{bold}{\encodingdefault}{\sfdefault}{bx}{n}

\title{LLMs as Adaptive Meta-Solvers: Strategy-Diverse RL for Industrial-Scale Optimization}

\author{%
\parbox{\textwidth}{\centering\vspace{6pt}
\textbf{Shihao~Zhang}$^{3*}$ \hspace{1em}
\textbf{Weiting~Liu}$^{2,4*}$ \hspace{1em}
\textbf{Siyu~Shao}$^{5}$ \hspace{1em}
\textbf{Yitian~Chen}$^{1\dagger}$  \\[2pt]
\textbf{Jianfeng~Feng}$^{4}$ \hspace{1em}
\textbf{Dongdong~Ge}$^{6}$ \hspace{1em}
\textbf{Yinyu~Ye}$^{6,7}$ \\[6pt]
\mdseries
$^1$Tokentide AI \quad
$^2$Alibaba Group \quad
$^3$East China Normal University \quad
$^4$Fudan University \\
$^5$The University of Hong Kong \quad
$^6$Shanghai Jiao Tong University \quad
$^7$Stanford University \\[6pt] 
}}

\usepackage{afterpage} 
\usepackage[table]{xcolor}
\usepackage{booktabs,multirow}
\definecolor{ourbg}{RGB}{232,240,252}   %
\usepackage[utf8]{inputenc} %
\usepackage[T1]{fontenc}    %
\usepackage{hyperref}       %
\usepackage{url}            %
\usepackage{booktabs}       %
\usepackage{amsfonts}       %
\usepackage{nicefrac}       %
\usepackage{microtype}      %
\usepackage{xcolor}         %
\usepackage{multirow}
\usepackage{graphicx}
\usepackage[most]{tcolorbox}
\usepackage{booktabs}
\usepackage{amsmath}
\usepackage{graphicx}
\usepackage{booktabs} 
\usepackage{enumitem}
\usepackage{subcaption} 
\usepackage{threeparttable}
\usepackage{tabularx}
\usepackage{titletoc}
\usepackage{wrapfig}

\usepackage{float}
\usepackage{xcolor}
\usepackage{fvextra}
\definecolor{algogreen}{HTML}{2E9E6B}
\definecolor{heurblue}{HTML}{3A6EA5}
\definecolor{optimindred}{HTML}{C0522A}

\usepackage{url}

\iclrfinalcopy %
\begin{document}

\maketitle
\lhead{Preprint.}

\begingroup
\renewcommand{\thefootnote}{}
\footnotetext{%
  \textsuperscript{*}Equal contribution.
  \textsuperscript{\ensuremath{\dagger}}Corresponding author:
  \texttt{chenyitian@tokentide.cn}%
}
\endgroup

\begin{abstract}
Scaling LLM-based optimization from textbook-scale instances to real-world, industrial tasks remains a critical open challenge. Existing approaches are predominantly evaluated on small, self-contained textual problems and often commit to a solver-integrated paradigm, limiting their ability to handle the scale and structural diversity of practical optimization workloads. In this work, we propose a practical framework for training open-source LLMs to tackle real-world, industrial-scale optimization. We first show empirically that solver-integrated reasoning, exact combinatorial algorithm, and heuristic search exhibit complementary strengths across different problem structures and scales. Motivated by this, we introduce \textbf{Strategy-Diverse Reinforcement Learning (SDRL)}, which trains LLMs as adaptive optimization meta-solvers. SDRL leverages this complementarity through a correctness-gated hierarchical diversity reward that promotes robust exploration across varying strategies and within each strategy, effectively preventing premature strategy collapse. We further introduce a mixed-format training scheme that jointly supports both self-contained textual problems and file-grounded instances. Across comprehensive evaluations, our framework outperforms existing fine-tuned methods and frontier models including DeepSeek-V4-Pro and GPT-5.5, both on average across benchmarks and on industrial-scale optimization tasks. We release our code and training data to facilitate future research \url{https://github.com/tokentideAI/LLMMetaSolver}.

\end{abstract}

\section{Introduction}
\label{sec:intro}
Optimization plays a fundamental role in decision-making across logistics, manufacturing, energy, finance, and many other real-world domains~\citep{singh2012overview,lu2007practical}. Recent advances in large language models (LLMs) have created new opportunities to automate this process, where LLMs translate natural-language descriptions into mathematical formulations and subsequently generate feasible solutions~\citep{huang2025orlm,astorga2024autoformulation}.
Despite this promise, existing methods are still largely evaluated on relatively small, \texttt{self-contained textual} problems and benchmarks where both problem descriptions and numerical parameters are embedded directly in the prompt~\citep{chen2026opt}. 
Real-world industrial optimization is substantially more challenging: practical instances are often \texttt{file-grounded}, involving thousands to millions of variables and constraints, while problem descriptions and instance-specific data are often distributed across external files rather than contained in a single prompt~\citep{li2026constructing,tang2026workspace,kong2026frontieror}. 
In such settings, optimization is no longer merely a one-shot autoformulation task, but an end-to-end problem-solving process. 
An effective optimization agent must interpret the problem context, access and analyze external data, identify the underlying structure, select an appropriate computational strategy, and construct an executable solution, closely mirroring the workflow of a human operations research (OR) expert~\citep{merrill2026terminal,fu2026opti}.

Real-world optimization also spans diverse problem structures and scales, demanding substantial strategic flexibility.
In practice, OR experts routinely choose among \texttt{mathematical formulation}~\citep{nemhauser1988integer}, \texttt{specialized algorithms} tailored for the problem~\citep{papadimitriou1998combinatorial}, and \texttt{heuristic search} methods~\citep{gendreau2010handbook} according to the characteristics of the problem. 
Relying exclusively on a single class of solution strategies can therefore be limiting: exact formulations are well suited to problems with compact mathematical representations and tractable scale, specialized algorithms can exploit problem-specific structure to achieve substantially greater efficiency, and heuristic methods often provide practical solutions when exact optimization becomes computationally prohibitive.
For example, large-scale TSP instances can make generic mathematical formulations increasingly costly in both representation and computation~\citep{reinelt1991tsplib}.

\begin{figure}[t]
    \centering
    \includegraphics[width=0.95\textwidth]{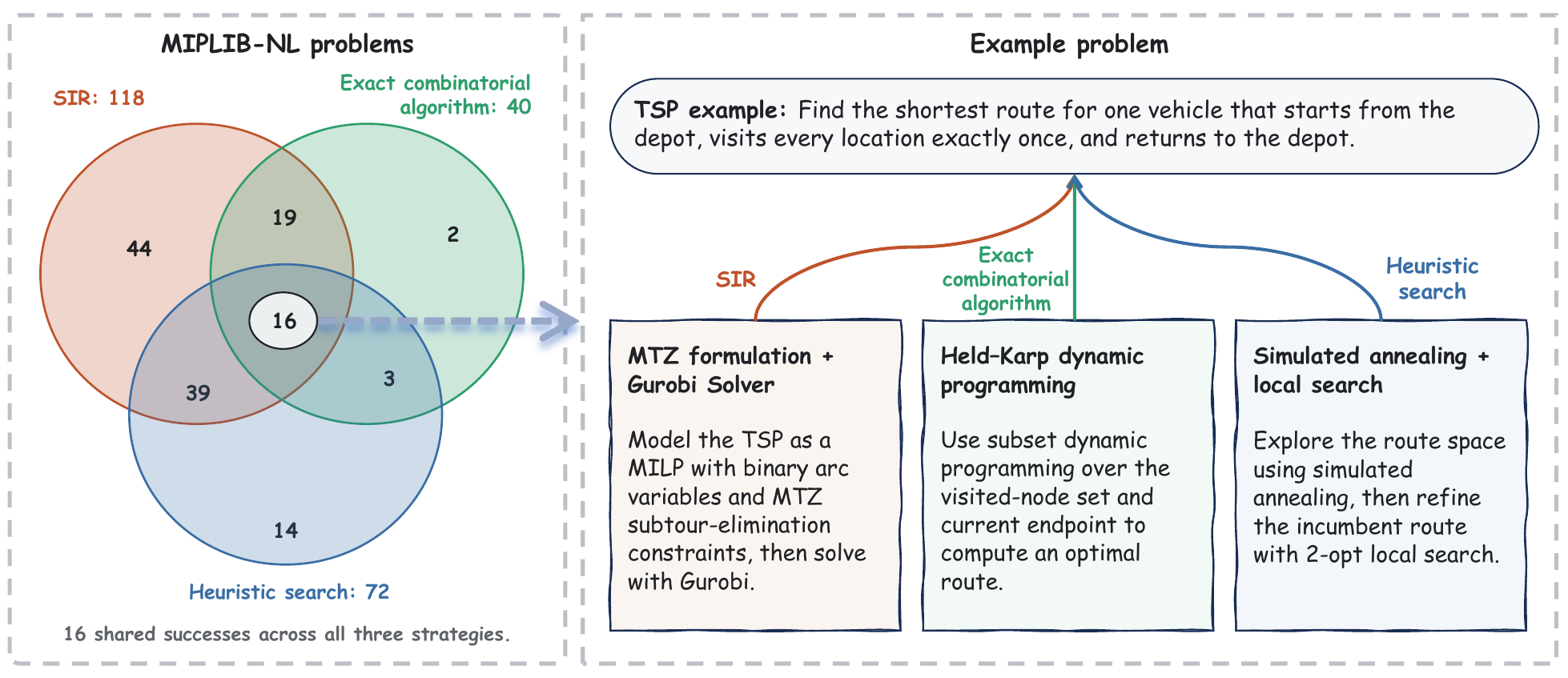}
    \caption{Motivating evidence for strategy complementarity. \textbf{Left:} Pass@8 success sets of DeepSeek-V4-Pro on MIPLIB-NL under three strategies. The partial overlap indicates that different strategies solve complementary subsets of instances. \textbf{Right:} a single TSP instance solved successfully by all three strategies, illustrating that one optimization problem can admit multiple valid computational solution paths.}
    \label{fig:overview}
\end{figure}

Current LLM-based optimization systems, however, rarely make this strategy choice adaptively. Instead, the solving paradigm is typically fixed by the system design.
Solver-integrated approaches typically focus on autoformulation for optimization problems, such as linear programming (LP) and mixed-integer linear programming (MILP), where LLMs translate natural-language descriptions into mathematical models and delegate computation to external solvers. In parallel, \emph{LLM-based automated algorithm design} searches directly over executable algorithms through iterative generation, evaluation, and code evolution~\citep{liu2024evolution,imajuku2026ale}. Such methods are particularly effective when optimizing reusable algorithms or heuristics for a fixed combinatorial problem family, such as TSP~\citep{reinelt1991tsplib} or CVRP~\citep{uchoa2017cvrp}. 
Despite their respective strengths, both paradigms typically commit to a predefined solution space rather than adaptively selecting among alternative computational strategies.
Figure~\ref{fig:overview} provides empirical evidence for this limitation on the challenging MIPLIB-NL benchmark~\citep{li2026constructing}. Under strategy-specific prompting, solver-based, algorithmic, and heuristic approaches yield only partially overlapping success sets, indicating substantial complementarity across strategies.
Their complementary performance motivates adaptive routing among multiple solution strategies for LLM-based optimization.

To this end, we propose a unified framework for training open-source LLMs~\citep{yang2025qwen3} as adaptive optimization meta-solvers capable of tackling real-world, industrial-scale problems.
At the algorithmic level, 
we introduce Strategy-Diverse Reinforcement Learning (SDRL). SDRL augments verifiable reinforcement learning with a correctness-gated, hierarchical diversity reward. This explicitly preserves multiple computational pathways—namely, solver-integrated reasoning, exact combinatorial algorithm, and heuristic search, thereby discouraging premature strategy collapse and promoting robust exploration both across and within these distinct solving paradigms.
At the systemic level, we develop a \emph{mixed-format training scheme} that jointly covers self-contained textual and file-grounded optimization problems, improving data and training efficiency while extending learning toward realistic industrial-scale optimization settings.

In summary, our main contributions are as follows:
\begin{enumerate}
\item \textbf{Empirical Analysis of Strategy Complementarity.}
We empirically demonstrate that, for LLM-based optimization, different problem classes and scales favor different solution paradigms, highlighting the limitations of relying on a single class of solution strategy.

\item \textbf{Strategy-Diverse Reinforcement Learning.}
We propose SDRL, a reinforcement learning framework that trains LLMs as optimization meta-solvers by encouraging diverse yet correct solution strategies across solver-integrated, algorithmic, and heuristic approaches.

\item \textbf{Mixed-Format Training and State-of-the-Art Performance.}
We introduce a mixed-format training scheme that combines self-contained textual problems with file-grounded problems with external structured data, improving data efficiency and supporting training on realistic industrial-scale optimization instances. Across the evaluated benchmarks, our 32B model achieves state-of-the-art Pass@1 performance, with larger gains in Pass@8.
\end{enumerate}

\section{Related Work}
\label{sec:relatedwork}
\textbf{Automated optimization modeling. }
    \textit{Automated optimization modeling} translates natural-language specifications into solver-ready mathematical formulations. Existing methods improve formulation quality through agent-based reasoning~\citep{ahmaditeshnizi2024optimus, xiao2024chain, zhang2024solving}, as well as search, knowledge augmentation, and verification~\citep{astorga2024autoformulation, liu2026optitree, kong2026alphaopt, liu2026opt}. Training-based approaches improve optimization modeling through supervised fine-tuning on curated or synthesized data~\citep{huang2025orlm, lu2025optmath, wu2025step, zhang2025optimind} and preference-based alignment~\citep{shu2025llmopt}. Further advances incorporate reinforcement learning with verifiable rewards~\citep{chen2026solver, xiao2026deepor, liu2026automated, zhao2026strategy, tang2025calm} and process-level supervision~\citep{zhou2026steporlm, wang2026or}. 
    However, existing methods are predominantly developed and evaluated on textbook-style tasks, leaving industrial-scale optimization largely underexplored.%

\textbf{Automated algorithm design. }
    Complementary to solver-based modeling, automated algorithm design seeks to improve computational efficiency by tailoring solution procedures to problem-specific structure. \textit{Automated exact algorithm design} constructs standalone, problem-specific algorithms for exact combinatorial optimization using techniques such as dynamic programming~\citep{zhou2025auto}, graph algorithms~\citep{nia2025evaluating}, and divide-and-conquer~\citep{huang2024effibench}, without relying on general-purpose mathematical programming solvers~\citep{wang2026formalize}. \textit{Automated heuristic design} focuses on generating and refining heuristics for optimization problems, commonly through evolutionary search over executable programs~\citep{romera2024mathematical, liu2024evolution}. Subsequent work improves the search itself through reflective and context-aware prompting~\citep{ye2024reevo, zhong2026hifo, bomer2025leveraging} and through diversity preservation, complementary heuristic sets, and multi-objective optimization~\citep{dat2025hsevo, liu2026eoh, yao2025multi}. Other work fine-tunes the language model via reinforcement learning, enabling it to co-evolve with heuristic search~\citep{huang2026calm}. These complementary approaches motivate instance-dependent paradigm selection. Rather than committing to a single solving paradigm, we learn a unified policy that selects among solver-integrated reasoning, exact combinatorial algorithm, and heuristic search based on the characteristics of each optimization instance.

\textbf{Exploration in complex reasoning.  }
    Another related topic is about preserving exploration during reinforcement learning for complex reasoning tasks. Recent studies have identified \textit{entropy collapse}, where reinforcement learning drives policies toward narrow reasoning patterns and reduces rollout diversity~\citep{yue2025does,cui2025entropy,wang2026beyond}. As training progresses, Pass@1 may continue to improve, while diminished exploration limits gains in Pass@k. Existing remedies operate at different granularities: token-level methods maintain exploration through entropy regularization or clipping~\citep{yu2026dapo}, while trajectory-level approaches encourage diverse reasoning paths through clustering, classification, or semantic-diversity rewards~\citep{hu2026rewarding,liang2026diverse,mishra2026sd,cao2026dpwriter}. Our formulation distinguishes itself from these approaches in two fundamental aspects. First, rather than measuring diversity at the token, textual, or general reasoning-trajectory level, we explicitly model algorithmic strategy diversity within the optimization domain. Second, instead of semantic similarity between generated texts, we ground diversity in execution behavior, distinguishing successful programs within a strategy by their runtime modes.

\section{Methodology}
\label{sec:methodology}

Motivated by the strategy complementarity in Figure~\ref{fig:overview}, we formulate LLM-based optimization as a {meta-solving} problem:
the model should first determine the proper solving paradigm before constructing a solution.
Mimicking the decision-making process of a human operations research expert, an LLM policy follows a structured three-stage trajectory: 
problem analysis, {strategy routing}, and {solution implementation}.
The policy first analyzes the problem's key characteristics, including the problem class (e.g., routing or scheduling), mathematical structure (e.g., continuous or discrete, linear or nonlinear), and estimated computational scale.
Based on this structural assessment, the policy identifies the most suitable computational paradigm among three candidate paradigms:
\begin{equation*}
\mathcal{S} =
\{\text{Solver-integrated reasoning},\
  \text{Exact combinatorial algorithm},\
  \text{Heuristic search}\}.
\end{equation*}
Here,  solver-integrated reasoning delegates a mathematical formulation to an external solver; exact combinatorial algorithm exploits problem-specific structure through specialized exact procedures; and heuristic search provides approximate solutions when exact methods are computationally expensive.
After routing, the policy generates the corresponding reasoning trace and executable solution. By explicitly separating strategy selection from solution construction, the meta-solver exposes the computational pathway as a learnable decision within the trajectory. This provides a natural foundation for reinforcement learning to jointly improve routing and solution quality under verifiable feedback.
 In the following subsections, we describe the training data construction and introduce \textbf{Strategy-Diverse Reinforcement Learning (SDRL)}, which augments verifiable rewards with a hierarchical diversity objective to promote diverse yet effective solution strategies.

\subsection{Training Data Construction}

\label{subsec:mixed_format}

\begin{figure*}[t]
    \centering
    \includegraphics[width=0.97\textwidth]{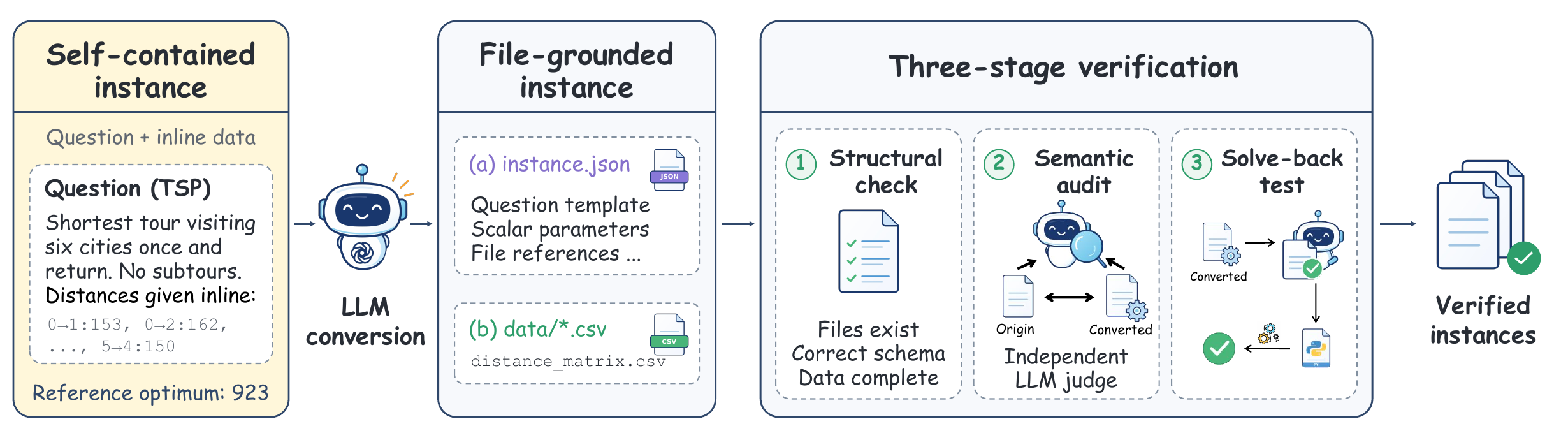}
    \caption{
    Overview of the file-grounded data construction pipeline.
    }
    \label{fig:data_construction}
\end{figure*}

Most existing NL-to-Opt training data~\citep{lu2025optmath} are self-contained textual examples, with both the problem specification and all instance-specific parameters embedded directly in the textual description. To equip the model with stronger agentic data-processing capabilities, we convert a subset of these examples into file-grounded instances using the pipeline illustrated in Figure~\ref{fig:data_construction}.

We represent each self-contained textual training example as
\begin{equation}
(p,o^\star),
\end{equation}
where $p$ denotes the natural-language problem specification and $o^\star$ is the reference optimal objective value. A conversion operator $\mathcal{T}$ transforms it into
\begin{equation}
\mathcal{T}(p,o^\star)
=
(\tilde{p},\mathcal{D},o^\star),
\end{equation}
with $\tilde{p}$ denoting the parameterized problem specification and
$\mathcal{D}=\{d_1,\ldots,d_m\}$ the associated external structured files. In practice, scalar parameters are stored in \texttt{instance.json}, while tabular data are stored in \texttt{data/*.csv}. The transformation preserves $o^\star$ while externalizing part of the instance-specific information.

Accordingly, the input optimization instance  can be uniformly formulated as
\begin{equation}
x=(p,\mathcal{D}),
\end{equation}
where $\mathcal{D}$ is optional: $\mathcal{D}=\emptyset$ for self-contained textual problems and $\mathcal{D}\neq\emptyset$ for file-grounded problems. For example, in a file-grounded TSP instance (Appendix~\ref{app:file_data}), $p$ specifies the task and optimization structure, including the objective, degree constraints, and MTZ subtour-elimination constraints, while the pairwise distance matrix is provided separately as a CSV file in $\mathcal{D}$.

\subsection{Strategy-Diverse Reinforcement Learning}
\label{subsec:sdrl}

\begin{figure*}[t]
    \centering
    \includegraphics[width=0.97\textwidth]{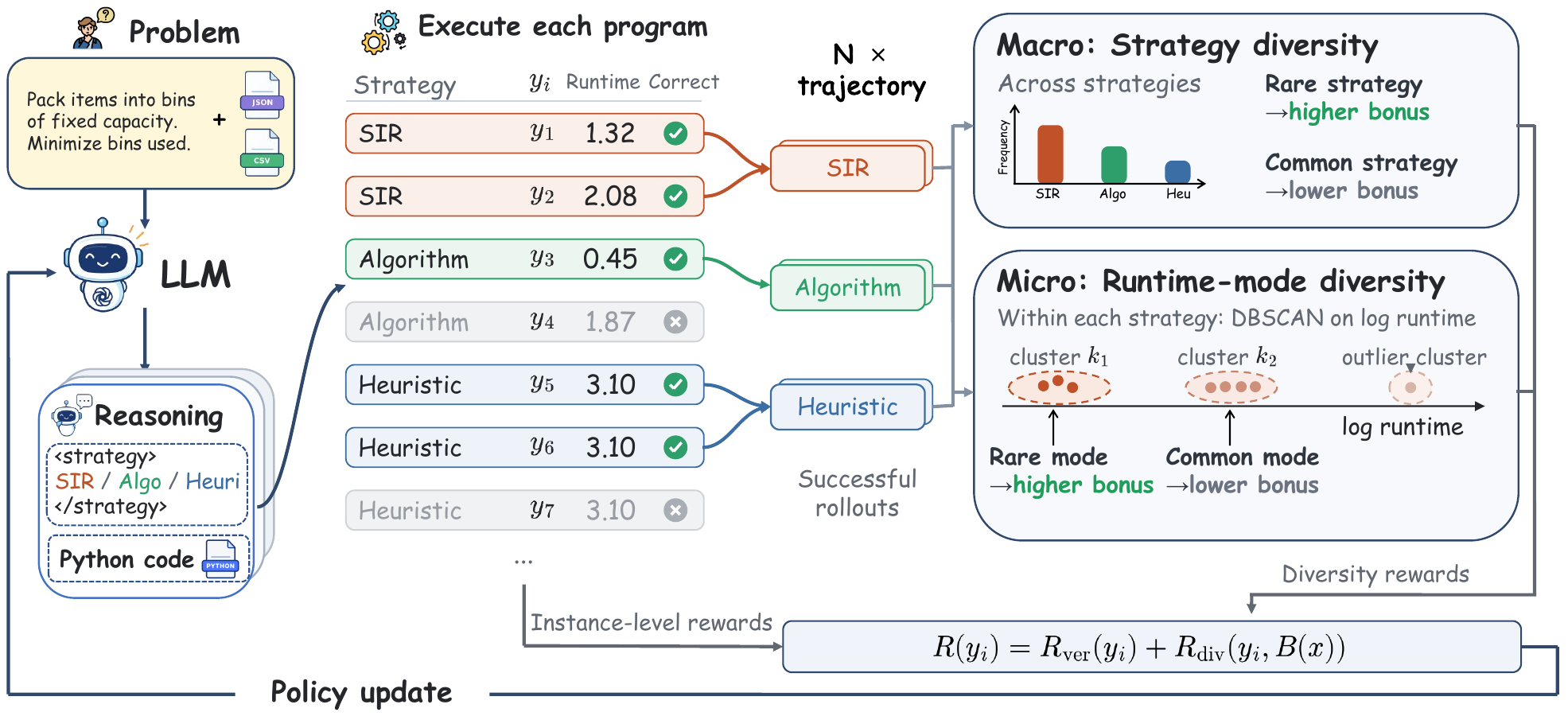}
    \caption{
    Overview of Strategy-Diverse Reinforcement Learning (SDRL).}
    
    \label{fig:framework}
\end{figure*}

We train the meta-solver policy $\pi_\theta$ using Group Relative Policy
Optimization (GRPO)~\citep{shao2024deepseekmath}.
Given an optimization instance $x$, the policy $\pi_\theta$ samples a group of $n$ rollouts
$G(x)=\{y_1,\ldots,y_n\}$ as shown in Figure~\ref{fig:framework}.
Guided by an expert-designed meta-prompt (Appendix~\ref{app:prompt}),
each rollout $y_i=(h_i, s_i,c_i)$ consists of a reasoning trace $h_i$, a selected strategy tag $s_i\in\mathcal{S}$, and an executable code-snippet
$c_i$. 
Each code block $c_i$ is then executed in the sandbox to obtain verifiable signals, including execution status, predicted objective value $\hat{o}_i$, and runtime $t_i$. 
These signals determine the instance-level verification reward $R_{\mathrm{ver}}(y_i)$ and thereby identify the set of successful rollouts
\begin{equation}
B(x)=\{y_i\in G(x): y_i \text{ is successful}\},
\label{eq:success_set}
\end{equation}
where success requires the generated program to execute correctly and its objective value to satisfy the prescribed evaluation tolerance.
Over $B(x)$, we compute a hierarchical diversity reward: the macro-level component favors underrepresented solution strategies, while the micro-level component promotes diverse runtime modes within each strategy. We then combine the verification and diversity terms to obtain the final rollout reward:
\begin{equation}
R(y_i)
=
R_{\mathrm{ver}}(y_i)
+
R_{\mathrm{div}}\bigl(y_i, B(x)\bigr),
\end{equation}
with diversity rewards applied only to successful trajectories.
Importantly, no ground-truth strategy labels are provided during training. Instead, the policy autonomously explores the strategy space, learning exclusively from the relative execution quality of its generated solutions. Because the downstream reasoning and code implementation are strictly conditioned on the upstream routing decision, the trajectory-level reward inherently supervises both stages.

Following GRPO, we compute a group-normalized advantage for each rollout and share it across all tokens in the response:
\begin{equation}
    \widehat{A}_i
    =
    \frac{
        R(y_i)-\bar{R}
    }{
        \operatorname{std}\!\bigl(R(y_1),\ldots,R(y_n)\bigr)+\delta
    },
    \qquad
    \bar{R}=\frac{1}{n}\sum_{j=1}^{n}R(y_j),
    \label{eq:group_advantage}
\end{equation}
where $\delta>0$ is a small constant for numerical stability.
The policy is then updated by minimizing the GRPO objective:
\begin{equation}
    \begin{aligned}
        \mathcal{L}_{\mathrm{GRPO}}(\theta)
        = -\,\mathbb{E}\Biggl[
            \frac{1}{n}\sum_{i=1}^{n}\frac{1}{|y_i|}\sum_{t=1}^{|y_i|}
            \min\bigl\{
                r_{i,t}(\theta)\,\widehat{A}_i,\;
                \operatorname{clip}\bigl(r_{i,t}(\theta),
                    1-\epsilon,1+\epsilon\bigr)
                \widehat{A}_i
            \bigr\}
        \Biggr] \\
        + \beta\,\mathbb{D}_{\mathrm{KL}}\bigl[\pi_\theta \,\|\, \pi_{\mathrm{ref}}\bigr],
    \end{aligned}
    \label{eq:grpo_loss}
\end{equation}
Here, $r_{i,t}(\theta)=
\frac{\pi_\theta(y_{i,t}\mid x,y_{i,<t})}
{\pi_{\theta_{\mathrm{old}}}(y_{i,t}\mid x,y_{i,<t})}$
is the token-level importance ratio, $\epsilon$ is the clipping threshold, and the KL term is estimated from samples against a frozen reference policy $\pi_{\mathrm{ref}}$ with weight
$\beta$.

Through iterative RLVR updates, the meta-solver policy $\pi_\theta$ progressively learns to select more suitable computational paradigms while improving implementation reliability and solution accuracy, resembling the decision process of a human operations research expert who adapts strategy based on execution feedback. 
In the next subsection, we formalize the reward framework that enables this joint learning, focusing specifically on our proposed hierarchical diversity reward.

\subsection{Reward Design and Training Scheme}
\label{subsec:div_reward_formulation}

\paragraph{Total Reward Framework.}
As illustrated in Figure~\ref{fig:framework},
the proposed SDRL framework combines two complementary sources of feedback for each rollout: an instance-level verification reward and a group-level strategy diversity reward. 
The verification reward evaluates whether an individual solution is valid, executable, and correct, whereas the diversity term differentiates among successful solutions according to the computational strategies they employ.

The final reward can be expressed as:
\begin{equation}
    R(y_i)
    =
    \underbrace{
        R_{\mathrm{fmt}}(y_i)
        +
        R_{\mathrm{exec}}(y_i)
        +
        R_{\mathrm{ans}}(y_i)
    }_{R_{\mathrm{ver}}(y_i)}
    +
    R_{\mathrm{div}}(y_i,B(x)),
    \label{eq:total_reward}
\end{equation}
where $R_{\mathrm{ver}}$ combines format validity, execution success, and objective-value correctness (Appendix~\ref{app:reward_func_design}) and $R_{\mathrm{div}}$ is the hierarchical diversity reward defined below.
The diversity term is correctness-gated:
\begin{equation}
R_{\mathrm{div}}(y_i,B(x))=0,
\qquad y_i\notin B(x).
\end{equation}
Consequently, diversity is rewarded only among successful trajectories, which encourages {strategic breadth} across paradigms while preserving {implementation proficiency} within each strategy.

\paragraph{Hierarchical Diversity Reward.}
The  diversity reward is constructed hierarchically at two levels. 
The \emph{macro-level} component promotes exploration across the three solution strategies, whereas the \emph{micro-level} component captures distinct runtime modes among successful rollouts within the same strategy. Both components use self-information to assign diversity credit at the trajectory level: successful rollouts receive larger bonuses when they adopt relatively uncommon strategies or runtime modes.

\textit{Macro-Level Strategy Diversity.}
Let $N=|B(x)|$ be the number of successful rollouts and $p_s$ the empirical frequency of strategy $s$ within $B(x)$. We define the macro-level reward as:

\begin{equation}
p_s
=
\frac{1}{N}
\sum_{y_j\in B(x)}
\mathbb{I}[s_j=s],
\qquad
R_{\mathrm{macro}}(y_i)
=
\begin{cases}
-\dfrac{\log p_{s_i}}{\log N},
& y_i\in B(x),\; N>1,\\[6pt]
0, & \text{otherwise}.
\end{cases}
\label{eq:macro_reward}
\end{equation}
Thus, successful rollouts that adopt less frequent strategies receive larger diversity bonuses. Since $p_{s_i}\geq 1/N$ for any observed strategy, normalization by $\log N$ bounds the reward in $[0,1]$.
To prevent reward hacking through strategy relabeling, we verify the declared strategy $s_i$ against the generated program using rule-based consistency checks. 
Rollouts whose declared strategy is inconsistent with their implementation (e.g., a trajectory labeled as heuristic search that invokes an exact solver) are excluded from $B(x)$ and receive no diversity reward (Appendix~\ref{app:reward_func_design}).

\textit{Micro-Level Runtime Diversity.}
High-level strategy labels alone do not capture all variation among successful solutions. Even within the same strategy, generated programs may exhibit distinct execution behaviors. We therefore use runtime patterns as a lightweight but effective signal of within-strategy diversity.
For each strategy $s$, we cluster the log-transformed runtimes of its $N_s$ successful rollouts using one-dimensional DBSCAN~\citep{ester1996density}. Let $\mathcal{K}_s$ denote the resulting runtime clusters, and let $m$ denote the minimum cluster-size parameter. Noise points are retained as a dedicated outlier cluster. 
For rollout $y_i$, let $k_i$ denote the cluster containing its runtime, let $n_{s_i,k_i}$ be the size of that cluster, and let $q_{s_i,k_i}=n_{s_i,k_i}/N_{s_i}$ be the corresponding cluster frequency. We define the micro-level reward as

\begin{equation}
R_{\mathrm{micro}}(y_i)
=
\begin{cases}
\operatorname{clip}\!\left(
\dfrac{-\log q_{s_i,k_i}}
{\log(N_{s_i}/m)},
0,1
\right),
&
y_i\in B(x),\;
|\mathcal{K}_{s_i}|\geq 2,
\\[8pt]
0,
&
\text{otherwise}.
\end{cases}
\label{eq:micro_reward}
\end{equation}

The reward is zero when only one runtime mode is observed within a strategy. For regular DBSCAN clusters, the minimum cluster size $m$ naturally bounds the normalized self-information. The clipping operation additionally prevents small outlier clusters from receiving disproportionately large diversity bonuses. We emphasize that this term measures \emph{runtime-mode rarity}, not slowness; Appendix D.4 confirms it does not induce slow programs.

Combining these two components yields the final diversity reward:
\begin{equation}
\label{eq:diversity_reward}
R_{\mathrm{div}}(y_i,B(x))
=
\mathbb{I}[y_i\in B(x)]
\left[
\lambda R_{\mathrm{macro}}(y_i)
+
(1-\lambda)R_{\mathrm{micro}}(y_i)
\right], 
\end{equation}

where $\lambda\in[0,1]$ controls the trade-off between cross-strategy and within-strategy diversity; we set $\lambda=0.7$ in the main experiments (sensitivity analysis in Appendix~\ref{app:lambda_sweep}). Compared with conventional RLVR, which primarily rewards correctness, SDRL further differentiates successful trajectories by their contribution to strategy and runtime-mode diversity, promoting broader exploration and mitigating premature strategy collapse.

\paragraph{Mixed-format Training.}

We train on the union of self-contained textual and file-grounded instances, uniformly shuffled, with the same response format and the reward in Eq.~\ref{eq:total_reward} for both types. For file-grounded instances, the training setup ensures that correctness can be achieved only by genuinely reading the data files. Each rollout executes in an isolated workspace that contains the instance's structured input files but not the reference optimum, and the prompt exposes only file paths and column names rather than table contents. Reading a file is not rewarded by itself; the rollout is scored by the same executable correctness criteria as a self-contained textual one.

\section{Experiments}
\label{sec:exp}
\subsection{Experimental Setups}

\textbf{Benchmarks.}
We evaluate on seven NL-to-Opt benchmarks:
NL4Opt~\citep{ramamonjison2023nl4opt},
MAMO-EasyLP and MAMO-ComplexLP~\citep{huang2025llms},
IndustryOR~\citep{huang2025orlm},
OptMATH-Bench~\citep{lu2025optmath},
OptiBench~\citep{yang2025optibench},
and MIPLIB-NL~\citep{li2026constructing}.
The first six use self-contained textual inputs.
 In contrast, MIPLIB-NL introduces file-grounded, industrial-scale optimization tasks that demand the ability to dynamically read and process external data files at runtime.
Benchmark details and instance counts are provided in Appendix~\ref{app:benchmark_details}.

\textbf{Baselines \& evaluation.}
We compare against Qwen3 base models~\citep{yang2025qwen3}, frontier LLMs, and representative fine-tuned NL-to-Opt models~\citep{huang2025orlm,lu2025optmath,chen2026solver,zhou2026steporlm}.
Following the strict evaluation protocol adopted in prior work~\citep{chen2026solver,lu2025optmath}, a prediction is considered correct if its relative objective error is below $10^{-6}$.
The full evaluation settings are provided in Appendix~\ref{app:eval_settings}.

\textbf{Training setup.}
For the main experiments, we train models initialized from Qwen3-4B-Instruct-2507 and Qwen3-32B~\citep{yang2025qwen3} under the mixed-format training scheme, while all ablation studies use Qwen3-4B-Instruct-2507 as the backbone. The complete training configuration is provided in Appendix~\ref{app:training_details}. We further distinguish between text-only and file-grounded training instances. The latter are constructed from the text-only instances by the file-grounded data pipeline described in Section~\ref{subsec:mixed_format}, and are combined with text-only instances to form our mixed-format training data.

\begin{table*}[t]
\centering
\caption{
Pass@1 accuracy across seven optimization benchmarks, with MIPLIB-NL representing a file-grounded, industrial-scale setting that requires agentic optimization modeling capabilities.
}
\label{tab:pass1_accuracy}
\setlength{\tabcolsep}{2pt}
\renewcommand{\arraystretch}{1.08}
\resizebox{\textwidth}{!}{%
\begin{tabular}{lcccccccc}
\toprule
\multirow{2}{*}{\textbf{Method}}
& \multicolumn{6}{c}{\textbf{Self-contained}}
& \multicolumn{1}{c}{\textbf{File-grounded}}
& \multirow{2}{*}{\textbf{Avg.}} \\
\cmidrule(lr){2-7}
\cmidrule(lr){8-8}
&
\textbf{NL4Opt}
& \textbf{MAMO-E}
& \textbf{MAMO-C}
& \textbf{Ind-OR}
& \textbf{OptMath}
& \textbf{OptiBench}
& \textbf{MIPLIB-NL}
& \\
\midrule
\multicolumn{9}{c}{\textit{Base Models}} \\
\midrule
\addlinespace[1pt]
Qwen3-4B-Instruct-2507
& 72.7 & 68.9 & 42.4 & 40.0 & 17.5 & 54.1 & 7.7 & 43.3 \\
Qwen3-32B
& 85.3 & 88.8 & 66.0 & 39.0 & 17.5 & 60.0 & 10.5 & 52.4 \\
\midrule

\multicolumn{9}{c}{\textit{Fine-tuned Models}} \\
\midrule
\addlinespace[1pt]
ORLM-Llama3-8B*
& 85.7 & 82.3 & 37.4 & 24.0 & 2.6 & 51.1 & 0.6 & 40.5 \\
OptMATH-Qwen2.5-7B*
& 94.7 & 86.5 & 51.2 & 20.0 & 24.4 & 57.9 & 0.0 & 47.8 \\
OptMATH-Qwen2.5-32B*
& 95.9 & 89.9 & 54.1 & 31.0 & 34.7 & 66.1 & 5.5 & 53.9 \\
SIRL-Qwen2.5-7B*
& \underline{96.3} & 91.7 & 51.7 & 33.0 & 30.5 & 58.0 & 0.5 & 51.7 \\
SIRL-Qwen2.5-32B*
& \textbf{98.0} & \underline{94.6} & 61.1 & 42.0 & \underline{45.8} & \underline{67.4} & 1.4 & 58.6 \\
StepORLM-Qwen3-8B
& 89.8 & 89.9 & 52.2 & 40.0 & 14.5 & 56.5 & 0.9 & 49.1 \\
\addlinespace[2pt]

\rowcolor{ourbg}
\textbf{SDRL-Qwen3-4B}
& 93.6 & 92.4 & \underline{79.3} & \underline{55.0} & 41.0 & 67.1 & \underline{23.6} & \underline{64.6} \\
\rowcolor{ourbg}
\textbf{SDRL-Qwen3-32B}
& \underline{96.3} & \textbf{96.0} & \textbf{81.8} & \textbf{56.0} & \textbf{59.0} & \textbf{69.1} & \textbf{30.0} & \textbf{69.7} \\
\midrule

\multicolumn{9}{c}{\textit{Frontier Models}} \\
\midrule
\addlinespace[1pt]
DeepSeek-V4-Pro
& 95.5 & 90.0 & 87.2 & 63.0 & 49.4 & 66.9 & 28.2 & 68.6 \\
Qwen3.5-397B-A17B
& 94.7 & 91.4 & 81.3 & 64.0 & 42.2 & 65.5 & 32.3 & 67.3 \\
GPT-5.5
& 95.5 & 93.8 & 92.1 & 67.0 & 40.4 & 69.6 & 29.1 & 69.6 \\
Claude-Opus-4.8
& 92.2 & 94.2 & 93.1 & 64.0 & 60.2 & 64.6 & 39.1 & 72.5 \\
Gemini-3.1-Pro-Preview
& 90.2 & 91.6 & 93.0 & 67.0 & 51.8 & 66.1 & 40.9 & 71.5 \\
\bottomrule

\end{tabular}%
}
{\raggedright Note: * denotes results from original or reproduced papers.\par} %
\end{table*}

\subsection{Main Results}

Table~\ref{tab:pass1_accuracy} presents the main results across the seven benchmarks. SDRL-Qwen3-4B outperforms all existing fine-tuned baselines, including larger 32B-parameter models such as OptMATH-Qwen2.5-32B~\citep{lu2025optmath} and SIRL-Qwen2.5-32B~\citep{chen2026solver}, while SDRL-Qwen3-32B achieves performance competitive with frontier LLMs. By explicitly incentivizing adaptive routing across solver-integrated, exact combinatorial algorithm and heuristic search solution strategies, SDRL establishes a new SOTA for open-source optimization modeling. 
The improvement is especially pronounced in the most challenging MIPLIB-NL benchmark~\citep{li2026constructing}, where instances are file-grounded, contain thousands of instance-specific parameters, and require stronger agentic capabilities for external-data processing, optimization modeling, and executable solution construction. Here, both models demonstrate agentic optimization modeling capabilities comparable to those of frontier LLMs.
Beyond the adaptive routing mechanism, we attribute an additional portion of these gains to our mixed-format training scheme. 
In the following subsection, we present ablation studies to quantify these contributing factors.

\subsection{Ablation Study of Hierarchical Diversity Reward}
\label{sec:ablation}
We next conduct an ablation study to isolate the individual contributions of the hierarchical diversity reward components.
Holding the text-only training instances and GRPO hyperparameters~\citep{shao2024deepseekmath} fixed, we compare five variants: the \textbf{Base model}, \textbf{SDRL w/o $R_{\mathrm{div}}$} (standard RL), \textbf{SDRL w/o $R_{\mathrm{micro}}$}, \textbf{SDRL w/o $R_{\mathrm{macro}}$}, and \textbf{Full SDRL}.
As shown in Table~\ref{tab:ablation}, standard RL alone substantially outperforms the base model, confirming the baseline benefit of verifiable accuracy signals. 
Building on this, 
the hierarchical diversity reward provides further gains, with the two components exhibiting clear complementarity.
The macro-level term explicitly encourages diverse strategic routing and solution pathways; this broadens exploration, which primarily elevates Pass@8. In contrast, the micro-level term promotes distinct implementations within each strategy, which improves single-sample reliability and mainly benefits Pass@1. Combining both in Full SDRL achieves the optimal balance of Pass@1 and Pass@8 and the largest gains on the challenging MIPLIB-NL benchmark.

\begin{table*}[htbp]
\centering
\caption{Ablation of the hierarchical diversity reward using text-only data.}
\label{tab:ablation}

\small
\renewcommand{\arraystretch}{1.05}

\begin{tabular*}{\textwidth}{
@{\extracolsep{\fill}}
lcccccccccc
@{}
}
\toprule
\multirow{2}{*}{\textbf{Method}}
& \multicolumn{2}{c}{MAMO-C}
& \multicolumn{2}{c}{IndustryOR}
& \multicolumn{2}{c}{OptMath}
& \multicolumn{2}{c}{MIPLIB-NL}
& \multicolumn{2}{c}{Average} \\
\cmidrule(lr){2-3}
\cmidrule(lr){4-5}
\cmidrule(lr){6-7}
\cmidrule(lr){8-9}
\cmidrule(lr){10-11}
& P@1 & P@8
& P@1 & P@8
& P@1 & P@8
& P@1 & P@8
& P@1 & P@8 \\
\midrule

Base model
& 42.4 & 76.9
& 40.0 & 62.0
& 17.5 & 36.8
& 7.7 & 19.1
& 26.9 & 48.7 \\

SDRL w/o $R_{\mathrm{div}}$
& 70.9 & 87.7
& 51.0 & 69.0
& 35.5 & 48.2
& 13.1 & 29.1
& 42.6 & 58.5 \\

SDRL w/o $R_{\mathrm{micro}}$
& 73.9 & 88.7
& 50.0 & \textbf{72.0}
& \textbf{37.4} & 53.6
& 14.6 & 31.8
& 44.0 & 61.5 \\

SDRL w/o $R_{\mathrm{macro}}$
& 75.9 & 88.2
& 52.0 & \textbf{72.0}
& 34.9 & 50.0
& 14.6 & 25.9
& 44.4 & 59.0 \\

Full SDRL
& \textbf{75.9} & \textbf{90.2}
& \textbf{53.0} & 71.0
& 36.7 & \textbf{54.8}
& \textbf{16.8} & \textbf{33.2}
& \textbf{45.6} & \textbf{62.3} \\

\bottomrule
\end{tabular*}
\end{table*}

\subsection{Ablation Study of Mixed-format Training}
\label{sec:mixed_format_ablation}

We examine the effect of the mixed-format training scheme by comparing
three settings: \textbf{Text-only}, \textbf{File-grounded}, and
\textbf{Mixed-format}. The file-grounded data are constructed from the
same underlying textual instances following the procedure in
Section~\ref{subsec:mixed_format}; Text-only and File-grounded therefore
differ only in input representation, not in content.

\begin{wraptable}[9]{r}{0.46\textwidth}
\centering
\small
\setlength{\tabcolsep}{4pt}
\caption{Ablation of the mixed-format training scheme.}
\label{tab:mixed_format_ablation}
\begin{tabular}{lcccc}
\toprule
\multirow{2}{*}{\textbf{Training Data}}
& \multicolumn{2}{c}{Text Avg.} & \multicolumn{2}{c}{MIPLIB-NL} \\
\cmidrule(lr){2-3}\cmidrule(lr){4-5}
& P@1 & P@8 & P@1 & P@8 \\
\midrule
Text-only     & 69.6 & \textbf{80.3} & 16.8 & 33.2 \\
File-grounded & 64.3 & 77.6 & 21.8 & \textbf{38.6} \\
Mixed-format  & \textbf{71.4} & 79.5 & \textbf{23.6} & 35.9 \\
\bottomrule
\end{tabular}
\vspace{-2pt}
\end{wraptable}
As shown in Table~\ref{tab:mixed_format_ablation}, the two formats exhibit clear complementarity. Text-only training attains the highest Pass@8 on self-contained textual benchmarks, whereas file-grounded training improves the model's ability to handle external-data instances, attaining the highest Pass@8 on MIPLIB-NL. However, each specialization comes at the cost of weaker performance outside its own domain. Mixed-format training achieves the best Pass@1 in both settings and provides the best balance overall.

\subsection{Training Dynamics}
\begin{figure*}[t]
    \centering
    \includegraphics[width=0.95\textwidth]{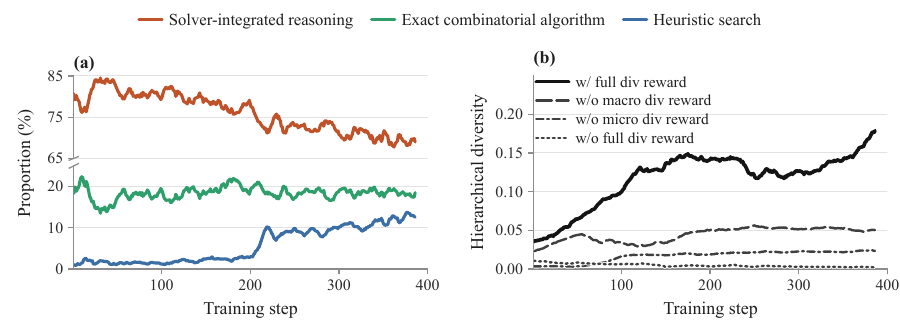}
    \vspace{-2mm}
    \caption{Training dynamics of SDRL.}
    \label{fig:training_dynamics}
    \vspace{-3mm}
\end{figure*}

We further analyze the training dynamics of SDRL to understand how the policy evolves during training. As shown in Figure~\ref{fig:training_dynamics}(a), the policy is initially dominated by solver-integrated reasoning (SIR), reflecting the strong solver-centric bias of existing LLM-based optimization methods. As training progresses, the hierarchical diversity reward encourages broader exploration across alternative computational pathways. The proportion of exact combinatorial algorithm remains relatively stable, which is consistent with their problem-specific nature and reliance on exploitable structure. Heuristic search, in contrast, increases steadily during training. This behavior is particularly informative given that the current evaluation protocol penalizes near-optimal solutions, suggesting that heuristic strategies may play an increasingly important role in scaling LLM-based optimization to industrial-scale settings. The bold curve in Figure~\ref{fig:training_dynamics}(b) tracks the hierarchical diversity score of Full SDRL throughout training, which consistently achieves the highest diversity among the four reward ablations.

\section{Conclusion and Limitations}
\label{sec:conclusion}

In this work, we present a framework for training LLMs to act as adaptive optimization meta-solvers. Strategy-Diverse Reinforcement Learning (SDRL) encourages exploration across solver-integrated reasoning, exact combinatorial algorithm, and heuristic search. Its hierarchical diversity reward gives diversity credit only to correct, executable solutions, favoring underrepresented strategies and runtime modes. This helps the model retain multiple viable approaches and reduces premature strategy collapse. We also introduce a mixed-format training scheme that supports both self-contained textual problems and file-grounded instances.
Across comprehensive evaluations, SDRL-Qwen3-4B outperforms all existing fine-tuned NL-to-Opt models in average Pass@1, while SDRL-Qwen3-32B exceeds DeepSeek-V4-Pro and GPT-5.5 while remaining competitive with other frontier LLMs. Notably, SDRL shows a clearer advantage when multiple solutions are sampled under Pass@k evaluation, suggesting broader coverage of viable strategies.
However, our current evaluation counts a solution as correct only when its objective value matches the reference within a strict tolerance. This may not fully reflect the value of heuristics that produce high-quality solutions at lower computational cost. Future work could develop evaluation metrics that consider both solution quality and computational cost, giving appropriate credit to efficient, near-optimal solutions.

\subsection*{AI use statement}

In this work, we used generative AI tools for generating and reformatting
training data: large language models convert self-contained optimization
instances into file-grounded ones and filter the converted instances, and
all retained instances are verified as documented in
Appendix~\ref{app:file_data}. Large language models are also evaluated as
baselines. Generative AI tools also assisted in writing and debugging
parts of the training and evaluation code, which the authors reviewed and
tested. Beyond these uses, we used generative AI tools only to polish
prose, improve clarity, and correct grammar; all such text was reviewed
and edited by the authors. The core ideas, methodology, experimental
design, and conclusions are the original work of the authors, who take
full responsibility for the final content of this work, including all
text, claims, and artifacts.

\subsection*{Ethics statement}

This work does not involve human subjects, personal data, or sensitive content. All training and evaluation data are derived from publicly available optimization benchmarks; the file-grounded training instances are machine-generated conversions of existing textual instances and are verified as described in Appendix~\ref{app:file_data}. Our method targets operations research problems such as routing, scheduling, and resource allocation, and we do not foresee direct negative societal impacts.

\subsection*{Reproducibility statement}

We have taken the following steps to make our results reproducible. The
complete system and user prompts used for training and evaluation, including
the strategy-routing meta prompt and the SIR prompt, are
given in Appendix~\ref{app:prompt}. The construction of the file-grounded
training data is described in Appendix~\ref{app:file_data}, including
the conversion prompt, a worked example, and the three-stage verification
protocol with its per-stage pass rates. The mixing ratio of the final training set is given in
Appendix~\ref{app:mixed_format_training}. All evaluation benchmarks and the
number of validated instances in each are listed in Appendix~\ref{app:benchmark_details}.
The full training configuration (backbone models, RL framework, and all
hyperparameters, including those of the diversity reward) is reported in
Appendix~\ref{app:training_details}, and the reward function, including the
rule-based strategy consistency checks, is specified in
Appendix~\ref{app:reward_func_design}. Decoding settings and the correctness
criterion for Pass@1 and Pass@8 are given in Appendix~\ref{app:eval_settings},
and the hardware used for runtime measurements in
Appendix~\ref{app:runtime}. Source code, training data, and evaluation
scripts will be released upon publication.

\bibliography{iclr2027_conference}
\bibliographystyle{iclr2027_conference}

\newpage
\appendix
\section*{APPENDIX}

\startcontents[appendix]               
\printcontents[appendix]{}{1}[2]{}   
\clearpage

\section{Prompt Templates}
\label{app:prompt}
\begin{tcolorbox}[
    enhanced,
    breakable,
    colback=gray!8!white,
    colframe=black,
    coltitle=white,
    colbacktitle=black,
    title=Meta Prompt Template,
    fonttitle=\bfseries,
    boxrule=0.8pt,
    arc=3pt,
    left=6pt,
    right=6pt,
    top=3pt,
    bottom=3pt,
    toptitle=2pt,
    bottomtitle=2pt,
    width=\textwidth,
    before skip=4pt,
    after skip=4pt
]
\begingroup
\scriptsize
\setlength{\parindent}{0pt}
\setlength{\parskip}{1pt}
\raggedright

\textbf{SYSTEM:}

You are an expert Operations Research Engineer and Algorithm Designer.
Your task is to analyze optimization problems, dynamically route them to the single most appropriate solution paradigm based on their mathematical characteristics and scale, and implement the solution in Python.

\textbf{Step 1: Problem Analysis}

Briefly analyze the problem based on the following dimensions:

\begin{itemize}[
    leftmargin=*,
    labelsep=4pt,
    itemsep=0pt,
    topsep=1pt,
    parsep=0pt,
    partopsep=0pt
]
    \item \textbf{Problem Type:}
    e.g., Routing, Scheduling, Assignment, Inventory, or Network Flow.

    \item \textbf{Mathematical Structure:}
    Determine whether the problem is continuous or discrete/combinatorial,
    linear or non-linear, and whether it exhibits optimal substructure.

    \item \textbf{Scale Estimation:}
    Approximate the number of variables and constraints, and determine whether
    the problem is small/medium-scale, for which exact solvers are manageable,
    or large-scale, for which combinatorial explosion may occur.
\end{itemize}

\textbf{Step 2: Strategy Routing (Choose ONE Paradigm)}

Based on the analysis, strictly select \textbf{ONE} of the following strategies:

\textcolor{optimindred}{\textbf{1. SIR}}

\textbf{Choose if:} The problem is a clearly defined Linear Programming (LP), Mixed-Integer Linear Programming (MILP), or Convex Quadratic problem, and the scale is manageable (small to medium).

\textbf{Mandatory:} You \textbf{MUST} use Gurobi through \texttt{gurobipy} for this strategy.

\textbf{Implementation:} Write a rigorous mathematical formulation using \texttt{gurobipy}. Start the code with:

{\ttfamily
\begin{tabular}{@{}l@{}}
import gurobipy as gp\\
from gurobipy import GRB
\end{tabular}
}

\textcolor{algogreen}{\textbf{2. Exact Combinatorial Algorithm}}

\textbf{Choose if:}
The problem exhibits optimal substructure, overlapping subproblems, or standard graph properties, such as knapsack, shortest path, or interval scheduling.

\textbf{Implementation:}
Use dynamic programming, greedy methods, divide-and-conquer, or backtracking.

\textcolor{heurblue}{\textbf{3. Heuristic Search}}

\textbf{Choose if:}
The problem is large-scale, highly non-linear, or a complex combinatorial optimization problem, such as a large-scale VRP or TSP instance, for which exact solvers are likely to time out.

\textbf{Implementation:}
Use a high-performance heuristic, such as ALNS, GRASP, iterated local search, or simulated annealing.

\textbf{Step 3: Output Format}

The response must follow the structure below.

First, output exactly one strategy tag on its own first line, using one of:

\textcolor{optimindred}{
\texttt{<strategy>SIR</strategy>}
}

\textcolor{algogreen}{
\texttt{<strategy>Exact Combinatorial Algorithm</strategy>}
}

\textcolor{heurblue}{
\texttt{<strategy>Heuristic Search</strategy>}
}

Then output the step-by-step reasoning, including the problem analysis and routing decision.

Finally, output the complete executable Python code in a single code block starting with \texttt{```python}.

\vspace{4pt}

\textbf{USER:}

Solve the following algorithm design and optimization problem:

\textcolor{blue}{\texttt{\{Question\}}}

Reason step by step to derive the logic before writing the script.
After thinking, explain your algorithmic strategy.
Finally, output the complete code block starting with \texttt{```python}.

Before the final output lines, you must assign \texttt{variables} to the computed
decision variable solution (prefer a list or a dictionary of variable values).
If the solution has no explicit decision-variable vector, set
\texttt{variables = []} as a fallback.

The script's final output lines must be exactly:

{\ttfamily
\begin{tabular}{@{}l@{}}
print("Decision variables:", variables)\\
print("Result:", objective\_value)
\end{tabular}
}

\noindent where \texttt{variables} is the optimal decision variable(s) and \texttt{objective\_value}
is the actual computed result variable.

\texttt{/no\_think}
\endgroup
\end{tcolorbox}

\begin{tcolorbox}[
    enhanced,
    breakable,
    colback=gray!8!white,
    colframe=black,
    coltitle=white,
    colbacktitle=black,
    title=SIR Prompt Template,
    fonttitle=\bfseries,
    boxrule=0.8pt,
    arc=3pt,
    left=6pt,
    right=6pt,
    top=3pt,
    bottom=3pt,
    toptitle=2pt,
    bottomtitle=2pt,
    width=\textwidth,
    before skip=4pt,
    after skip=4pt
]
\begingroup
\scriptsize
\setlength{\parindent}{0pt}
\setlength{\parskip}{1pt}
\raggedright

\textbf{SYSTEM:}

You are a helpful assistant with expertise in mathematical modeling and the Gurobi solver. When the user provides an Operations Research (OR) problem, you must:

\begin{enumerate}[
    leftmargin=*,
    labelsep=4pt,
    itemsep=0pt,
    topsep=1pt,
    parsep=0pt,
    partopsep=0pt
]
    \item Carefully analyze the problem and clearly define:
    \begin{itemize}[
        leftmargin=*,
        labelsep=4pt,
        itemsep=0pt,
        topsep=1pt,
        parsep=0pt,
        partopsep=0pt
    ]
        \item Decision variables
        \item Objective function
        \item Constraints
    \end{itemize}
    Carefully determine whether each decision variable is integer or continuous.
    \item Build a complete mathematical model.
    \item Provide Gurobi Python code that formulates and solves the model.
\end{enumerate}

Ensure correctness, clarity, and professional presentation.

\vspace{4pt}

\textbf{USER:}

Here is the given optimization problem:

\textcolor{blue}{\texttt{\{Question\}}}

Reason step by step to derive the modeling process before generating the \texttt{gurobipy} code. When you respond, first think carefully. After thinking, output the mathematical model. Finally, output a code block beginning with:

{\ttfamily
\begin{tabular}{@{}l@{}}
\texttt{```python}\\
import gurobipy as gp\\
from gurobipy import GRB
\end{tabular}
}

The script's final output line must be:
{\ttfamily
\begin{tabular}{@{}l@{}}
print("Result:", objective\_value)
\end{tabular}
}

where \texttt{objective\_value} is the actual computed objective value.

\texttt{/no\_think}

\endgroup
\end{tcolorbox}

\section{Datasets}
\subsection{Construction of File-Grounded Training Data}
\label{app:file_data}
This section describes how file-grounded training instances are constructed
from self-contained textual ones (Section~\ref{app:file_data_conversion}) and how every converted instance is verified before being used for training
(Section~\ref{app:file_data_verification}).

\subsubsection{Conversion}
\label{app:file_data_conversion}

File-grounded training instances are derived from the OptMATH training set by
separating the optimization specification from instance-specific data. For
each self-contained question, DeepSeek-V4-Pro (temperature 0.1, JSON output)
produces three components: a parameterized \texttt{abstract\_problem}, a
dictionary of scalar \texttt{parameters}, and a set of structured tables that
are serialized as external CSV files. Values already stored in a table are not
duplicated in the parameter dictionary. The original optimal objective value is
retained as \texttt{optimal\_value} and serves as the supervision target, so
the conversion changes the information-access pattern of an instance but not
the optimization task itself.

\paragraph{Conversion Prompt.}
The following prompt is used to convert the original self-contained
optimization instances into the file-grounded format.

\begin{tcolorbox}[
    enhanced,
    breakable,
    lines before break=4,
    colback=gray!8!white,
    colframe=black,
    coltitle=white,
    colbacktitle=black,
    title=File-Grounded Data Construction Prompt,
    fonttitle=\bfseries,
    boxrule=0.8pt,
    arc=3pt,
    left=6pt,
    right=6pt,
    top=3pt,
    bottom=3pt,
    toptitle=2pt,
    bottomtitle=2pt,
    width=\textwidth,
    before skip=4pt,
    after skip=4pt
]
\begingroup
\scriptsize
\setlength{\parindent}{0pt}
\setlength{\parskip}{1pt}
\raggedright

You are an expert in Operations Research and Data Engineering. Your task is to
convert a natural language optimization problem into a structured JSON format
that enforces problem-data separation---decoupling the abstract natural
language formulation from instance data---without missing any information.

\vspace{4pt}

\textbf{INSTRUCTIONS:}

\begin{enumerate}[
    leftmargin=*,
    labelsep=4pt,
    itemsep=2pt,
    topsep=1pt,
    parsep=0pt,
    partopsep=0pt
]

\item \textbf{abstract\_problem}: Translate the original problem description
into a clear, concise, and parameterized natural language formulation. You must
strictly preserve all original constraints, objectives, and logical
relationships without omitting any details. Replace instance-specific scalar
numbers (e.g., limits, costs, capacities) with variable placeholders
(e.g., \texttt{\{num\_jobs\}}, \texttt{\{budget\_limit\}}). However, exercise
expert judgment to retain certain explicit numbers (e.g., structural constants,
specific indices, dimensions, or binary states).

\item \textbf{parameters}: Extract the scalar values corresponding to the
placeholders defined in your \texttt{abstract\_problem} as key-value pairs.
Do not include placeholders if their values are explicitly contained within
the extracted tables data (to prevent data redundancy).

\item \textbf{tables}: Extract tabular data. For each table, provide a key,
\texttt{description}, \texttt{header} (list of strings), and
\texttt{data\_matrix} (list of lists). CRITICAL: The description must
explicitly define the exact meaning of each column header included in the
table. Furthermore, it must clearly explain how missing, inapplicable, or empty
values are represented within the \texttt{data\_matrix}.

\end{enumerate}

\vspace{4pt}

\textbf{EXAMPLE OUTPUT FORMAT:}

\begin{lstlisting}
{
  "abstract_problem": "Optimize a scheduling problem with {n_jobs} jobs
  under a budget limit of {budget_limit}, subject to precedence constraints.",

  "parameters": {
    "n_jobs": 3,
    "budget_limit": 100
  },

  "tables": {
    "jobs": {
      "description": "A table containing each job's duration and cost.",
      "header": ["job_id", "duration", "cost"],
      "data_matrix": [
        [1, 2, 10],
        [2, 3, 20],
        [3, 4, 15]
      ]
    },

    "precedence": {
      "description": "Each row specifies a predecessor-successor relationship.",
      "header": ["predecessor_id", "successor_id"],
      "data_matrix": [
        [1, 3]
      ]
    }
  }
}
\end{lstlisting}

\vspace{4pt}

\textbf{OUTPUT FORMAT CONSTRAINT:}

You must output ONLY a valid JSON object matching the exact schema shown in the
example above.

\endgroup
\end{tcolorbox}

\paragraph{Conversion Example.}
A concrete OptMATH example is shown below to illustrate how a self-contained
optimization problem is converted into a file-grounded instance.

\begin{tcolorbox}[
    enhanced,
    breakable,
    colback=gray!8!white,
    colframe=black,
    coltitle=white,
    colbacktitle=black,
    title=Conversion Example: Six-City Asymmetric TSP,
    fonttitle=\bfseries,
    boxrule=0.8pt,
    arc=3pt,
    left=6pt,
    right=6pt,
    top=4pt,
    bottom=4pt,
    toptitle=2pt,
    bottomtitle=2pt,
    width=\textwidth,
    before skip=4pt,
    after skip=4pt
]
\begingroup
\scriptsize
\setlength{\parindent}{0pt}
\setlength{\parskip}{2pt}
\raggedright

\textbf{Original Self-Contained Question.}

The original OptMATH instance is a six-city asymmetric traveling salesperson
problem. A delivery service must visit City 0--City 5 exactly once and return
to the starting city while minimizing the total travel distance. The question
directly specifies all pairwise travel distances between the six cities,
together with the in-degree and out-degree constraints required for a valid
tour. It further introduces auxiliary order variables and MTZ-style constraints
to prevent disconnected subtours. Thus, both the optimization specification
and all instance-specific numerical data are contained directly in the
natural-language input.

\vspace{5pt}
\hrule
\vspace{5pt}

\textbf{File-Grounded Instance Example}

After conversion, the optimization specification and instance-specific data are stored separately:

\vspace{2pt}

{\ttfamily
\begin{tabular}{@{}l@{}}
instance\_00000/ \\
\quad instance.json \\
\quad data/ \\
\qquad distance\_matrix.csv
\end{tabular}
}

\vspace{6pt}

\textbf{\texttt{instance.json}}

The instance description contains the parameterized optimization problem,
scalar parameters, references to external files, and the original supervision
target:

\vspace{2pt}

\begin{tabular}{@{}p{0.24\linewidth}p{0.70\linewidth}@{}}
\texttt{abstract\_problem}
&
Parameterized TSP specification containing the objective, degree constraints,
and MTZ subtour-elimination constraints, while referring to pairwise distances
as externally provided data.
\\[3pt]

\texttt{parameters}
&
\texttt{n = 6}, \texttt{n\_minus\_one = 5},
\texttt{big\_M = 5}.
\\[3pt]

\texttt{files}
&
\texttt{distance\_matrix}
$\rightarrow$
\texttt{./data/distance\_matrix.csv}.
\\[3pt]

\texttt{optimal\_value}
&
\texttt{923.0}.
\end{tabular}

\vspace{6pt}

\textbf{\texttt{data/distance\_matrix.csv}}

The complete asymmetric distance matrix, originally embedded in the
natural-language question, is moved to an external CSV file:

\vspace{3pt}

\begin{center}
\setlength{\tabcolsep}{4.5pt}
\renewcommand{\arraystretch}{1.0}
\begin{tabular}{c|rrrrrr}
\toprule
\textbf{from\_city}
& \textbf{to\_0}
& \textbf{to\_1}
& \textbf{to\_2}
& \textbf{to\_3}
& \textbf{to\_4}
& \textbf{to\_5} \\
\midrule
0 & 0   & 153 & 162 & 167 & 166 & 165 \\
1 & 157 & 0   & 157 & 152 & 160 & 163 \\
2 & 153 & 158 & 0   & 152 & 155 & 164 \\
3 & 170 & 166 & 169 & 0   & 161 & 161 \\
4 & 150 & 161 & 168 & 170 & 0   & 156 \\
5 & 154 & 164 & 157 & 150 & 150 & 0   \\
\bottomrule
\end{tabular}
\end{center}

\vspace{5pt}
\hrule
\vspace{5pt}

\textbf{Preserved Supervision.}

The original optimal objective value is retained as
\texttt{optimal\_value = 923.0}. One optimal tour is

\[
0 \rightarrow 1 \rightarrow 2 \rightarrow 3
\rightarrow 5 \rightarrow 4 \rightarrow 0,
\]

with total distance

\[
153 + 157 + 152 + 161 + 150 + 150 = 923.
\]

The conversion therefore preserves the optimization task and supervision
signal, while moving instance-specific numerical data from the textual prompt
to external files that must be accessed at runtime.

\endgroup
\end{tcolorbox}

\subsubsection{Verification}
\label{app:file_data_verification}

Because the rewriting is performed by a model while the label is carried over
unchanged, a dropped constraint or a truncated table would produce an instance
whose supervision signal no longer matches its specification, without any
error being raised. We therefore verify every converted instance in three stages of increasing
cost (Table~\ref{tab:file_data_verification}), and an instance is retained
for training only if it passes all three. Stage~1 is deterministic and
discards instances with structural or numerical defects; Stages~2 and~3
require model calls and are applied to every Stage-1 survivor.
Stage~2 uses Claude-Sonnet-4.6 as
the judge and Stage~3 uses Claude-Opus-4.8 as the solver; the stronger model is
reserved for solving so that a Stage-3 failure reflects a broken instance
rather than a weak solver. Both belong to neither the family of the conversion
model (DeepSeek-V4-Pro) nor that of the trained Qwen3 backbones, so that a
shared prior cannot repair an ambiguous conversion and the trained model
cannot influence which instances are kept.

\begin{table}[h]
\centering
\caption{Verification of converted file-grounded training instances. Each
stage is applied to all instances that pass the preceding stage, and pass
rates are conditional on the preceding stage. Only instances passing all
three stages are used for training.}
\label{tab:file_data_verification}
\small
\setlength{\tabcolsep}{6pt}
\begin{tabular}{llr}
\toprule
\textbf{Stage} & \textbf{Check} & \textbf{Pass rate} \\
\midrule
1 & Structural and numerical integrity (deterministic)       & 88.2\% \\
2 & Source--conversion consistency (Claude-Sonnet-4.6)        & 86.2\% \\
3 & Solve-back reproduces the optimal value (Claude-Opus-4.8) & 82.1\% \\
\bottomrule
\end{tabular}
\end{table}

\paragraph{Stage 1: Structural and Numerical Integrity.}
Deterministic checks that require no model call. An instance is discarded if
it produces no external table; if a referenced CSV file is missing, empty, or
has rows whose width differs from the header; if a placeholder in the problem
text resolves to neither a parameter nor a table column; if more than 10\% of
the numerical values in the source question cannot be found in the parameters
or tables, or the tables contain values absent from the source; if a declared
count such as \texttt{num\_constraints} is not realised by the row count of
any table; or if the optimal value appears in the problem text. The last check
matters because a leaked label would make Stage~3 trivially pass.

\paragraph{Stage 2: Source--Conversion Consistency.}
An independent judge (Claude-Sonnet-4.6) is shown the source question and the
converted instance and decides whether they describe the same optimization problem,
checking the objective direction, the decision variables, every constraint,
index ranges, and the meaning and units of each column. It returns
\texttt{consistent}, \texttt{inconsistent}, or \texttt{uncertain} together
with a confidence; an instance passes if it is judged \texttt{consistent}
with confidence at least 0.7. The judge receives the system prompt below,
followed by a user message that contains the source question inside
\texttt{<ORIGINAL\_PROBLEM>} tags and the converted instance inside
\texttt{<CONVERTED\_PROBLEM>} tags. The converted instance is rendered as a
JSON object with the \texttt{abstract\_problem}, the \texttt{parameters}, and,
for each table, its path, description, column names, row count, and a six-row
head/tail preview. The optimal value is never included.

\begin{tcolorbox}[
    enhanced,
    breakable,
    colback=gray!8!white,
    colframe=black,
    coltitle=white,
    colbacktitle=black,
    title=Stage-2 Consistency Judge Prompt,
    fonttitle=\bfseries,
    boxrule=0.8pt,
    arc=3pt,
    left=6pt,
    right=6pt,
    top=3pt,
    bottom=3pt,
    toptitle=2pt,
    bottomtitle=2pt,
    width=\textwidth,
    before skip=4pt,
    after skip=4pt
]
\begingroup
\scriptsize
\setlength{\parindent}{0pt}
\setlength{\parskip}{2pt}
\raggedright

\textbf{System.}

You are a meticulous auditor of Operations Research problem conversions.

The ORIGINAL problem is a self-contained natural-language optimization
question. The CONVERTED problem intentionally moves instance data into
external CSV files and may replace concrete values with placeholders. Moving
values to files and rewriting the prose is expected and is not, by itself, an
inconsistency.

Decide whether the converted representation still describes the same problem.
Check, in particular:

\begin{itemize}[leftmargin=*, labelsep=4pt, itemsep=1pt, topsep=1pt, parsep=0pt]
\item minimization versus maximization and the exact objective meaning;
\item what is being selected/decided, variable domains, and required quantities;
\item every important constraint, coverage/flow/assignment relationship, and
      index range;
\item whether each CSV column and its data type/meaning matches the source data;
\item whether a table is missing, has the wrong orientation, wrong units, or
      the wrong number of entities;
\item whether a condition was added, removed, or materially changed.
\end{itemize}

Do not use an optimal value or any answer/output as evidence. Do not penalize
the converted problem merely because numeric values are in CSV files instead
of prose. If the supplied table preview is insufficient to establish a claim,
use \texttt{uncertain} rather than inventing facts. Be conservative: an
omitted or contradictory objective/constraint is \texttt{inconsistent}.

Return ONLY one JSON object with this schema:

{\ttfamily
\begin{tabular}{@{}l@{}}
\{ \\
\quad "verdict": "consistent" | "inconsistent" | "uncertain", \\
\quad "confidence": number between 0 and 1, \\
\quad "summary": "one or two sentence explanation", \\
\quad "issues": [ \\
\qquad \{"category": "objective|variables|constraints|data|indices|other", \\
\qquad\ "description": "specific issue, or why no issue was found"\} \\
\quad ] \\
\}
\end{tabular}
}

\vspace{4pt}
\hrule
\vspace{4pt}

\textbf{User.}

{\ttfamily
\begin{tabular}{@{}l@{}}
<ORIGINAL\_PROBLEM> \\
\{source question\} \\
</ORIGINAL\_PROBLEM> \\[3pt]
<CONVERTED\_PROBLEM> \\
\{ "abstract\_problem": ..., "parameters": \{...\}, \\
\ \ "files": \{ \{key\}: \{ "path": ..., "description": ..., \\
\qquad\qquad\qquad\ "table": \{ "columns": [...], "row\_count": N, \\
\qquad\qquad\qquad\qquad\qquad\ "preview\_rows": [...] \} \} \} \} \\
</CONVERTED\_PROBLEM> \\[3pt]
Compare the two representations and return the required JSON object.
\end{tabular}
}

\endgroup
\end{tcolorbox}

\paragraph{Stage 3: Solve-Back.}
Given only the converted instance, Claude-Opus-4.8 writes a program that reads
the CSV files at runtime and solves the problem. The program is executed in an
isolated workspace into which the CSV files are materialized, with a
120-second limit. An instance passes if one of $k=3$ attempts (one greedy,
two sampled) reproduces the original optimal value under the relative
tolerance of $10^{-6}$ used throughout our evaluation. This stage deliberately
reuses the training prompt: the strategy-routing system prompt and the output
contract of Appendix~\ref{app:prompt}, with the \texttt{/no\_think} directive
removed because it is specific to the Qwen backbones. Only the user-message
body differs from a self-contained instance, and it is the same body that
file-grounded instances receive during training. Because prompt, executor,
and tolerance coincide with the training setup, an instance that fails this
stage is one on which the policy could never obtain a correctness reward. The
template below is instantiated with the six-city TSP example of
Section~\ref{app:file_data_conversion}; note that the prompt exposes file
paths, column names, and row counts but never the table contents.

\begin{tcolorbox}[
    enhanced,
    breakable,
    colback=gray!8!white,
    colframe=black,
    coltitle=white,
    colbacktitle=black,
    title=Stage-3 Solve-Back Prompt (File-Grounded Question Body),
    fonttitle=\bfseries,
    boxrule=0.8pt,
    arc=3pt,
    left=6pt,
    right=6pt,
    top=3pt,
    bottom=3pt,
    toptitle=2pt,
    bottomtitle=2pt,
    width=\textwidth,
    before skip=4pt,
    after skip=4pt
]
\begingroup
\scriptsize
\setlength{\parindent}{0pt}
\setlength{\parskip}{2pt}
\raggedright

\textbf{System.} Strategy-routing prompt of Appendix~\ref{app:prompt}.

\vspace{4pt}
\hrule
\vspace{4pt}

\textbf{User.}

Solve the following algorithm design and optimization problem:

Problem name: instance\_00000

We are tasked with finding the shortest Hamiltonian cycle for a traveling
salesperson problem involving 6 cities. The cities are indexed from 0 to 5.
[\ldots] The distances between all pairs of cities are provided in a distance
matrix. The goal is to determine the optimal sequence of visits that minimizes
total travel distance while respecting these constraints.

Instance parameters:\\
\texttt{\{ "big\_M": 5, "n": 6, "n\_minus\_one": 5 \}}

Instance data is stored in external CSV files.\\
Your generated Python code will run from an isolated instance workspace.\\
Load the required table data at runtime from the exact relative paths below;
do not hardcode CSV rows or table values in the code.

\begin{itemize}[leftmargin=*, labelsep=4pt, itemsep=1pt, topsep=1pt, parsep=0pt]
\item distance\_matrix (./data/distance\_matrix.csv)\\
      Exact CSV columns: ["from\_city", "to\_0", "to\_1", "to\_2", "to\_3",
      "to\_4", "to\_5"]\\
      Data rows (excluding header): 6\\
      Description: A table representing the asymmetric distance matrix between
      the 6 cities. The header contains `from\_city' followed by the
      destination cities `to\_0', \ldots, `to\_5'. Each row corresponds to a
      starting city, and the columns represent the distance to the destination
      city. The diagonal distances are represented as 0, but these self-loops
      are not allowed in the tour. The matrix is not necessarily symmetric.
\end{itemize}

Determine the optimal objective value for this instance.

Reason step by step to derive the logic before writing the script. After
thinking, explain your algorithmic strategy. Finally, output the complete code
block starting with \texttt{```python}. Before the final output lines, you
must assign \texttt{variables} to the computed decision variable solution
(prefer a list or a dictionary of variable values). If the solution has no
explicit decision-variable vector, set \texttt{variables = []} as a fallback.
The script's final output lines must be exactly:\\
\texttt{print("Decision variables:", variables)}\\
\texttt{print("Result:", objective\_value)}\\
where \texttt{variables} is the optimal decision variable(s) and
\texttt{objective\_value} is the actual computed result variable.

\endgroup
\end{tcolorbox}
 
\subsection{Benchmarks}
\label{app:benchmark_details}

We evaluate SDRL on seven optimization modeling benchmarks spanning a range
of problem structures, difficulty levels, and application domains, from
classical linear programming instances to industrial-scale combinatorial
optimization problems. Table~\ref{tab:benchmark_stats} summarizes the number
of validated problem instances in each benchmark used in our experiments.

\begin{itemize}
    \item \textbf{NL4Opt}~\citep{ramamonjison2023nl4opt}. A benchmark of
    natural language descriptions of linear programming problems, focusing
    on the translation from textual problem statements into formal LP
    models. Problems span classical operations research settings such as
    resource allocation and production planning.

    \item \textbf{MAMO-EasyLP / MAMO-ComplexLP}~\citep{huang2025llms}. A
    modeling-oriented benchmark designed to assess formulation correctness
    rather than solution accuracy alone. MAMO is split into an EasyLP
    subset, consisting of standard LP formulations, and a ComplexLP subset,
    which introduces additional structural complexity such as nested
    conditions and hierarchical constraints.

    \item \textbf{IndustryOR}~\citep{huang2025orlm}. A real-world industrial
    benchmark comprising optimization problems drawn from manufacturing,
    logistics, finance, and energy domains. Problems in IndustryOR are
    annotated by difficulty level and frequently exhibit implicit or
    non-standard constraint structures that deviate from textbook
    formulations.

    \item \textbf{OptMATH-Bench}~\citep{lu2025optmath}. A benchmark designed
    to cover a wide range of optimization paradigms beyond standard linear
    programming, including mixed-integer, nonlinear, and combinatorial
    formulations, with an emphasis on diverse problem topologies.

    \item \textbf{OptiBench}~\citep{yang2025optibench}. A large-scale
    benchmark constructed via reverse Socratic synthesis, in which
    optimization demonstrations are transformed into natural language
    problem descriptions, providing high-quality intermediate reasoning
    structure alongside the final formulations.

    \item \textbf{MIPLIB-NL}~\citep{li2026constructing}. A benchmark
    constructed from industrial-scale mixed-integer programming instances,
    featuring large numbers of variables and constraints that pose
    significant challenges for both exact and heuristic solving approaches.
    Three instances in the original release ship with empty data files
    and cannot be instantiated; we exclude them from all experiments in
    this paper and report results on the remaining 220 instances.
\end{itemize}

\begin{table}[htbp]
\centering
\caption{Number of validated problem instances used in each evaluation
benchmark. For MIPLIB-NL, three instances with empty data files are
excluded from the original 223.}
\label{tab:benchmark_stats}
\begin{tabular}{lc}
\toprule
\textbf{Benchmark} & \textbf{\# Instances} \\
\midrule
NL4Opt         & 245 \\
MAMO-EasyLP    & 642 \\
MAMO-ComplexLP & 203 \\
IndustryOR     & 100 \\
OptMATH-Bench  & 166 \\
OptiBench      & 605 \\
MIPLIB-NL      & 220 \\
\bottomrule
\end{tabular}
\end{table}

\section{Details of experiments}
\subsection{Training Parameters}
\label{app:training_details}
We used Qwen3-4B-Instruct-2507 and Qwen3-32B~\citep{yang2025qwen3} as the
backbone models. The reinforcement learning stage was implemented based on
the veRL framework~\citep{sheng2024hybridflow} using Group Relative Policy
Optimization (GRPO)~\citep{shao2024deepseekmath}. Following
DAPO~\citep{yu2026dapo}, we adopt asymmetric clipping with
$\epsilon_{\mathrm{low}}=0.20$ and $\epsilon_{\mathrm{high}}=0.28$ in
place of the symmetric threshold in Eq.~\ref{eq:grpo_loss}, which raises
the ceiling on low-probability tokens and mitigates entropy collapse. We extended the reward computation pipeline to incorporate executable correctness verification and our hierarchical diversity reward, which considers both macro-level strategy diversity and micro-level runtime diversity (Section~\ref{subsec:div_reward_formulation}). Unless otherwise specified, the two backbone models followed the same training configuration. All experiments were conducted on eight NVIDIA H200 GPUs.

The key hyperparameters used for reinforcement learning are summarized in
Table~\ref{tab:training_parameters}.

\begin{table}[htbp]
    \centering
    \small
    \caption{Training parameters.}
    \label{tab:training_parameters}
    \begin{tabular}{lll}
        \toprule
        \textbf{Type} & \textbf{Parameter} & \textbf{Value} \\
        \midrule

        \textbf{Model}
        & Backbone (small) & Qwen3-4B-Instruct-2507 \\
        & Backbone (large) & Qwen3-32B \\

        \midrule
        \textbf{Algorithm}
        & Advantage estimator & GRPO \\
        & Training steps & 400 \\

        \midrule
        \textbf{Data}
        & Batch size & 64 \\
        & Learning rate & $1\times10^{-6}$ \\
        & Max prompt length & 5{,}120 \\
        & Max response length & 16{,}384 \\

        \midrule
        \textbf{Actor/Rollout}
        & KL loss type & \texttt{low\_var\_kl} \\
        & KL loss coefficient & 0.001 \\
        & Rollout number & 16 \\
        & PPO mini-batch size & 16 \\
        & PPO micro-batch size per GPU & 2 \\
        & Clip ratio low & 0.20 \\
        & Clip ratio high & 0.28 \\

        \midrule
        \textbf{Diversity reward}
        & Macro/micro weight $\lambda$ & 0.7 \\
        & DBSCAN neighborhood $\epsilon$ (log-runtime) & 0.25 \\
        & Minimum cluster size $m$ & 2 \\

        \bottomrule
    \end{tabular}
\end{table}

\subsection{Mixed-Format Training Details}
\label{app:mixed_format_training}

For the main SDRL training, we jointly use about 17K self-contained textual
instances and 3K file-grounded instances, shuffled uniformly. In self-contained instances, all information required to construct the optimization problem is provided directly in the prompt. In file-grounded instances, part of the instance-specific information remains in external structured files and can be accessed by the generated program at runtime.

The same response format and reward function are used for both input formats.
No additional reward is introduced for accessing external files; file-grounded
responses are evaluated according to whether the generated program executes
successfully and produces the correct objective value.

Unless otherwise specified, the hierarchical diversity reward ablation in
Section~\ref{sec:ablation} is trained without file-grounded instances in order
to isolate the effect of the diversity reward from that of mixed-format
training.

\subsection{Reward Function Design}
\label{app:reward_func_design}

Given a generated response $y_i$, we define its reward as the sum of
instance-level quality rewards and a group-level strategy diversity reward:
\begin{equation}
    R(y_i)
    =
    R_{\mathrm{fmt}}(y_i)
    + R_{\mathrm{exec}}(y_i)
    + R_{\mathrm{ans}}(y_i)
    + R_{\mathrm{div}}(y_i,B(x)).
\end{equation}

Here, $R_{\mathrm{fmt}}$, $R_{\mathrm{exec}}$, and $R_{\mathrm{ans}}$
measure the format validity, execution success, and objective-value
correctness of an individual response, respectively. $R_{\mathrm{div}}$ is
computed at the rollout-group level following the hierarchical macro/micro
formulation described in Section~\ref{subsec:div_reward_formulation}, and is
assigned only to correct rollouts.

\paragraph{1. Format reward.}

The format reward jointly enforces the code-block convention and the
strategy-routing protocol. We define two binary indicators:
$\mathbb{I}_{\mathrm{code}}(y_i)$, which indicates whether $y_i$ contains a
fenced Python code block, and $\mathbb{I}_{\mathrm{tag}}(y_i)$, which
indicates whether $y_i$ begins with exactly one valid strategy tag of the
form \texttt{<strategy>}$s_i$\texttt{</strategy>}, where

\begin{equation}
    s_i \in
    \{
    \texttt{SIR},\;
    \texttt{Exact Combinatorial Algorithm},\;
    \texttt{Heuristic Search}
    \}.
\end{equation}

The format reward is then defined as the average of the two indicators,

\begin{equation}
    R_{\mathrm{fmt}}(y_i)
    =
    \frac{1}{2}\,\mathbb{I}_{\mathrm{code}}(y_i)
    +
    \frac{1}{2}\,\mathbb{I}_{\mathrm{tag}}(y_i),
\end{equation}

so that a response receives full credit only when both requirements are
satisfied, and partial credit when only one of the two is met.

\paragraph{2. Execution reward.}

We extract and execute the Python code from the response. The execution
reward penalizes syntax errors, runtime errors, and invalid programs:

\begin{equation}
R_{\mathrm{exec}}(y_i) =
\begin{cases}
1, & \text{if } \mathrm{Exec}(y_i) = \mathrm{Success}, \\
0, & \text{otherwise}.
\end{cases}
\end{equation}

For file-grounded instances, the generated program can additionally access
the associated external files during execution.

\paragraph{3. Answer reward.}

Let $\hat{o}_i$ be the objective value obtained from executing the generated
program, and let $o^\star$ be the ground-truth optimal value. We compute the
relative error as

\begin{equation}
e_i = \frac{|\hat{o}_i-o^\star|}{|o^\star|+\epsilon},
\end{equation}

where $\epsilon = 10^{-6}$ is a small constant for numerical stability. The
answer reward is defined as

\begin{equation}
R_{\mathrm{ans}}(y_i) =
\begin{cases}
3, & e_i < \tau_{\mathrm{strict}}, \\
1, & \tau_{\mathrm{strict}} \le e_i < \tau_{\mathrm{relaxed}}, \\
0, & e_i \ge \tau_{\mathrm{relaxed}},
\end{cases}
\end{equation}

where $\tau_{\mathrm{strict}}=10^{-6}$ and
$\tau_{\mathrm{relaxed}}=10^{-3}$ in our experiments.

\paragraph{4. Diversity reward.}

$R_{\mathrm{div}}(y_i,B(x))$ follows the hierarchical macro/micro diversity
formulation introduced in Section~\ref{subsec:div_reward_formulation}.
The macro-level component assigns larger diversity bonuses to correct
rollouts using underrepresented solution strategies, while the micro-level
component rewards rare runtime modes within each strategy. Both components
use normalized self-information to assign trajectory-level diversity credit.

Since the macro-level component favors correct rollouts whose declared
strategy is underrepresented in the group, a policy could in principle
exploit it by attaching a rare strategy tag (e.g., Heuristic Search) to a
program that actually solves the instance with an exact solver. To prevent
this form of reward hacking, the declared strategy $s_i$ is verified
against the content of the response with lightweight rule-based checks
before it is used for diversity computation. We collect three types of
evidence from each response: SIR evidence, detected only from the
extracted code block via solver-specific imports and API calls (e.g.,
\texttt{import gurobipy}, \texttt{gp.Model}, \texttt{addVars},
\texttt{optimize}), so that merely mentioning a solver in the explanation
does not count; Exact Combinatorial Algorithm evidence, detected from both
the explanation and the code via keywords and library usage characteristic
of such algorithms (e.g., dynamic programming, greedy, shortest path, max
flow, \texttt{heapq}, \texttt{networkx}); and Heuristic Search evidence,
detected analogously via keywords such as local search, simulated
annealing, tabu, genetic algorithm, and 2-opt. A declared strategy is
accepted only if it is supported by its own evidence and not contradicted
by stronger evidence for another strategy. In particular, SIR requires at
least one solver call, Heuristic Search requires at least one heuristic
indicator and no solver call, and Exact Combinatorial Algorithm requires
no solver call and no more heuristic than algorithmic evidence.

A response that fails this check is treated as having no valid strategy:
it receives $R_{\mathrm{div}}(y_i,B(x))=0$ and is excluded from the
strategy distribution over which the macro-level self-information is
computed, so that mislabeled rollouts cannot inflate or dilute the
diversity credit of other rollouts in the same group. The instance-level
rewards $R_{\mathrm{fmt}}$, $R_{\mathrm{exec}}$, and $R_{\mathrm{ans}}$
are unaffected by this check. As a result, the diversity bonus rewards
genuinely different solution strategies rather than superficial changes to
the strategy tag.

\subsection{Evaluation Sampling Settings}
\label{app:eval_settings}
For Pass@1 evaluation, each problem instance is solved with a single
generation under temperature $0.5$ and top-$p$ $0.95$. For Pass@$K$ evaluation
(Section~\ref{app:passk_results}), we independently sample $K=8$ responses
per instance under temperature $1.0$ and top-$p$ $0.95$, and an instance is
considered solved if at least one of the $K$ responses is correct. For both
settings, we use a repetition penalty of $1.05$ and disable top-$k$
truncation (top-$k=-1$).

A generated response is considered correct if the extracted program executes
successfully and its predicted objective value $\hat{o}$ satisfies the
following relative tolerance criterion:
\begin{equation}
\frac{|\hat{o}-o^\star|}{|o^\star|+\epsilon} < 10^{-6},
\end{equation}
where $o^\star$ is the reference objective value and $\epsilon=10^{-6}$ is a
small constant introduced to avoid division by zero when $o^\star=0$,
consistent with the tolerance used throughout our main results.

\subsection{Detailed Pass@1 and Pass@8 Results}
\label{app:passk_results}

Table~\ref{tab:appendix_passk} reports the detailed Pass@1 and Pass@8 results
of the Base Model and SDRL under the Qwen3-4B-Instruct-2507 and Qwen3-32B
backbones across all seven evaluation benchmarks. All models are evaluated
using the same decoding configuration. Pass@1 measures single-sample
performance, while Pass@8 measures whether at least one correct executable
solution is obtained among eight sampled responses.

\begin{table*}[htbp]
\centering
\caption{
Detailed Pass@1 and Pass@8 results (\%) of the Base Model and SDRL
using Qwen3-4B-Instruct-2507 and Qwen3-32B across seven benchmarks.
}
\label{tab:appendix_passk}
\setlength{\tabcolsep}{3.5pt}
\renewcommand{\arraystretch}{1.05}

\resizebox{\textwidth}{!}{
\begin{tabular}{llcccccccc}
\toprule
\textbf{Method}
& \textbf{Metric}
& \textbf{NL4Opt}
& \textbf{MAMO-E}
& \textbf{MAMO-C}
& \textbf{IndustryOR}
& \textbf{OptMATH}
& \textbf{OptiBench}
& \textbf{MIPLIB-NL}
& \textbf{Avg.} \\
\midrule

\multirow{2}{*}{Qwen3-4B-Instruct}
& P@1
& 72.7 & 68.9 & 42.4 & 40.0
& 17.5 & 54.1 & 7.7
& 43.3 \\

& P@8
& 93.5 & 89.6 & 76.9 & 62.0
& 36.8 & 70.1 & 19.1
& 64.0 \\

\multirow{2}{*}{SDRL-Qwen3-4B}
& P@1
& \textbf{93.6}
& \textbf{92.4}
& \textbf{79.3}
& \textbf{55.0}
& \textbf{41.0}
& \textbf{67.1}
& \textbf{23.6}
& \textbf{64.6} \\

& P@8
& \textbf{95.1}
& \textbf{96.3}
& \textbf{88.2}
& \textbf{70.0}
& \textbf{54.2}
& \textbf{73.2}
& \textbf{35.9}
& \textbf{73.3} \\

\midrule

\multirow{2}{*}{Qwen3-32B}
& P@1
& 85.3 & 88.8 & 66.0 & 39.0
& 17.5 & 60.0 & 10.5
& 52.4 \\

& P@8
& 95.5 & 96.3 & 86.2 & 67.0
& 42.2 & 70.3 & 27.7
& 69.3 \\

\multirow{2}{*}{SDRL-Qwen3-32B}
& P@1
& \textbf{96.3}
& \textbf{96.0}
& \textbf{81.8}
& \textbf{56.0}
& \textbf{59.0}
& \textbf{69.1}
& \textbf{30.0}
& \textbf{69.7} \\

& P@8
& \textbf{98.0}
& \textbf{97.3}
& \textbf{88.3}
& \textbf{69.0}
& \textbf{75.3}
& \textbf{73.1}
& \textbf{45.0}
& \textbf{78.0} \\

\bottomrule
\end{tabular}}
\end{table*}

Overall, SDRL consistently improves both Pass@1 and Pass@8 across the two
model scales, showing gains in both single-sample reliability and
repeated-sampling coverage.

\section{Further Analysis}

\subsection{Effect of Training Data Size}
\label{app:data_scaling}

\begin{figure}[p]
    \centering
    \begin{subfigure}[b]{\textwidth}
        \centering
        \includegraphics[width=0.8\textwidth]{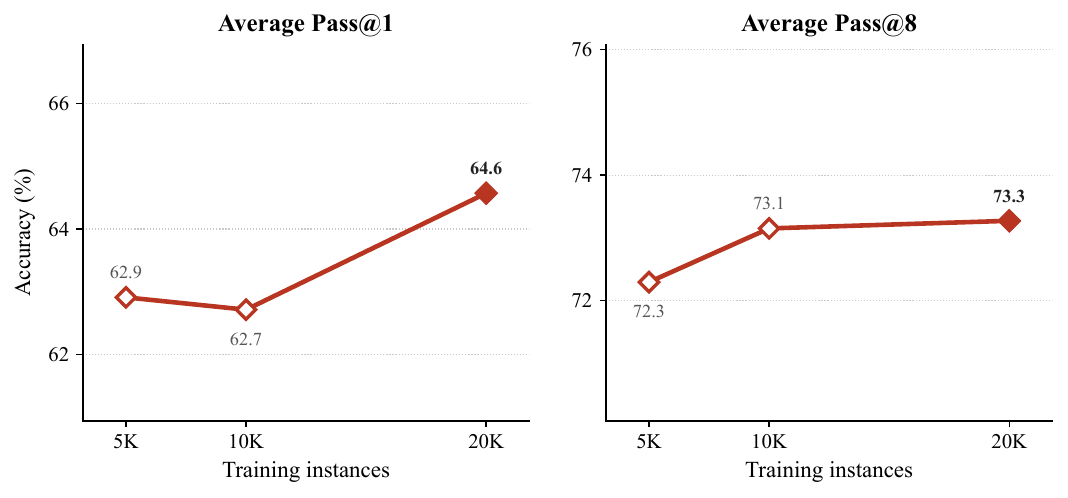}
        \caption{Average over the seven benchmarks (left: Pass@1, right: Pass@8).}
        \label{fig:data_scaling_avg}
    \end{subfigure}
    \vspace{6pt}
    \begin{subfigure}[b]{\textwidth}
        \centering
        \includegraphics[width=\textwidth]{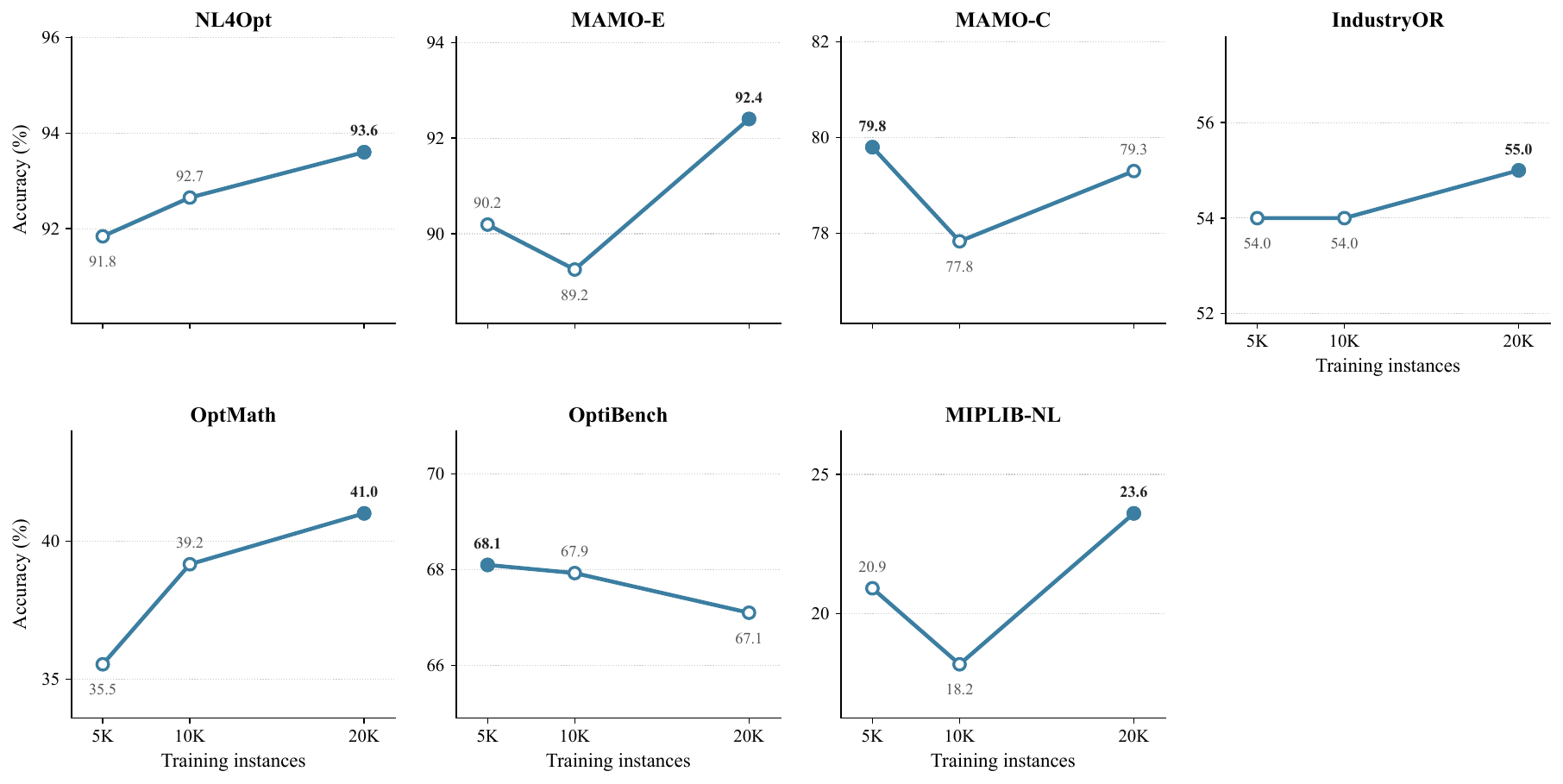}
        \caption{Per-benchmark Pass@1.}
        \label{fig:data_scaling_pass1}
    \end{subfigure}
    \vspace{6pt}
    \begin{subfigure}[b]{\textwidth}
        \centering
        \includegraphics[width=\textwidth]{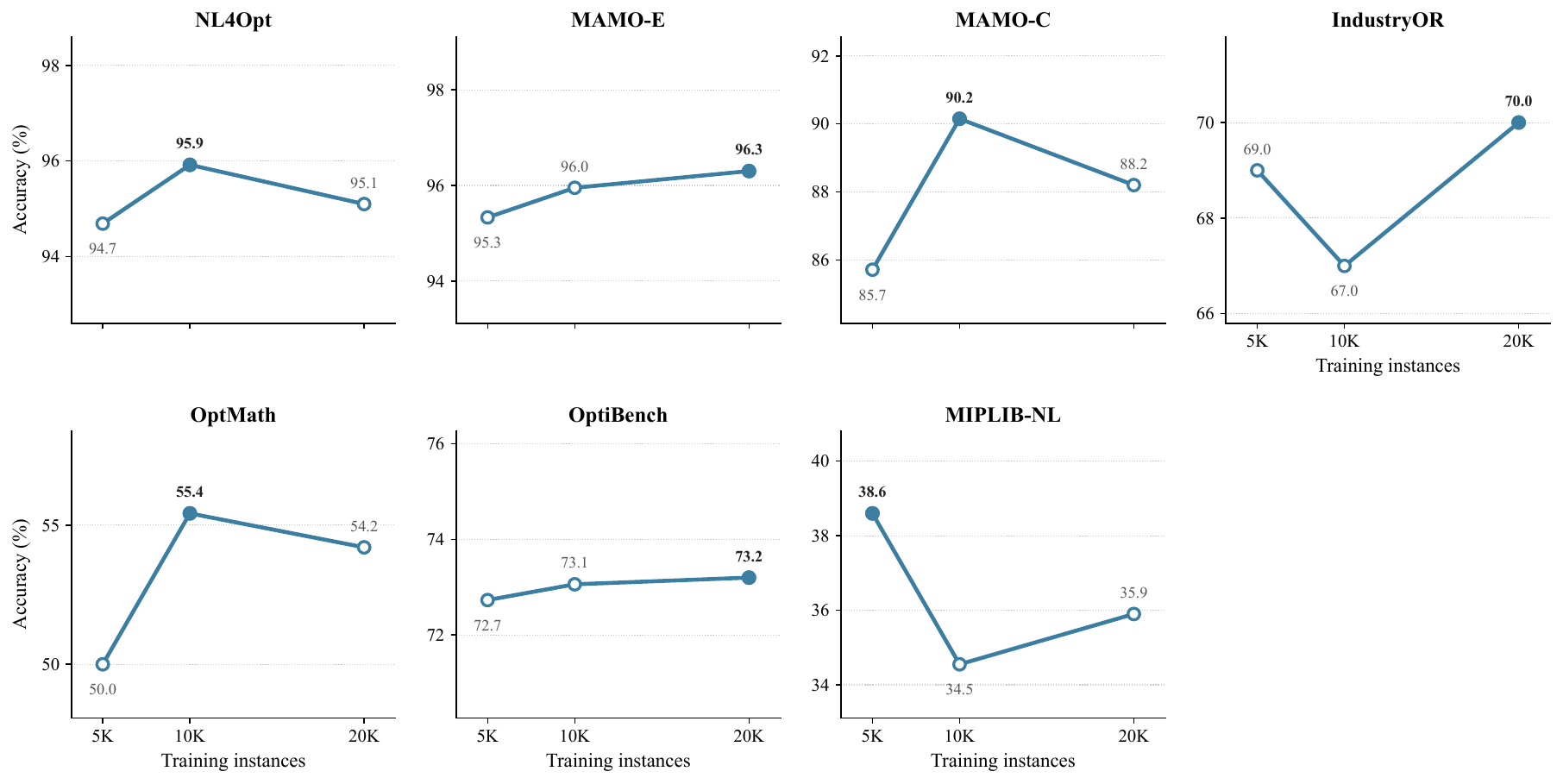}
        \caption{Per-benchmark Pass@8.}
        \label{fig:data_scaling_pass8}
    \end{subfigure}
    \caption{Effect of training data size for SDRL (Qwen3-4B-Instruct-2507)
    trained with 5K, 10K, and 20K mixed-format instances. Each panel uses
    its own y-axis range; filled markers denote the best setting.}
    \label{fig:data_scaling}
\end{figure}
We examine whether SDRL benefits from more distinct training instances
under the mixed-format training setting. SDRL is trained with 5K, 10K, and 20K instances drawn from the
same mixed-format data pool, using Qwen3-4B-Instruct-2507 as the backbone.
All three runs use exactly the configuration of
Table~\ref{tab:training_parameters}: the same reward formulation and
diversity-reward hyperparameters, the same ratio of self-contained to
file-grounded instances, the same GRPO hyperparameters, and the same number
of training steps (400) with the same batch size. The runs therefore consume
an identical optimization budget and differ only in the number of distinct
instances available, with smaller sets being revisited more often. The 20K
setting corresponds to the SDRL model in the main results. All models are
evaluated with the decoding configuration of
Section~\ref{app:eval_settings}.

Figure~\ref{fig:data_scaling} reports the results: Figure~\ref{fig:data_scaling_avg}
shows the average accuracy over the seven benchmarks, and
Figures~\ref{fig:data_scaling_pass1} and~\ref{fig:data_scaling_pass8} the
per-benchmark Pass@1 and Pass@8. Average Pass@1 is essentially flat between
5K and 10K (62.9 and 62.7) and improves to 64.6 at 20K, while average
Pass@8 rises from 72.3 to 73.1 between 5K and 10K and then plateaus. The
gains are concentrated on the harder benchmarks such as OptMATH-Bench and
MIPLIB-NL. Notably, even with only 5K instances, SDRL already surpasses all
fine-tuned baselines in Table~\ref{tab:pass1_accuracy} on average Pass@1,
indicating that the diversity-driven exploration is data-efficient rather
than dependent on scale. We nonetheless adopt 20K instances as the default
training set size, as it gives the best overall accuracy.

\subsection{Sensitivity to the Macro/Micro Weight $\lambda$}
\label{app:lambda_sweep}

The weight $\lambda$ in Eq.~\ref{eq:diversity_reward} balances
cross-strategy exploration ($R_{\mathrm{macro}}$) against within-strategy
runtime diversity ($R_{\mathrm{micro}}$). We sweep
$\lambda\in\{0,0.1,0.3,0.5,0.7,0.9,1.0\}$ on Qwen3-4B-Instruct-2507 with the
text-only training data and the configuration of
Table~\ref{tab:training_parameters}, changing only $\lambda$. The endpoints
$\lambda=0$ and $\lambda=1$ correspond to the micro-only and macro-only
variants of Table~\ref{tab:ablation}, and $\lambda=0.7$ is the default used
throughout the paper. Table~\ref{tab:lambda_sweep} reports Pass@1 and Pass@8
on the four benchmarks of the ablation study.

\begin{table}[htbp]
\centering
\caption{Effect of the macro/micro weight $\lambda$ on
Qwen3-4B-Instruct-2507 (text-only training). $\lambda=0$ uses only
$R_{\mathrm{micro}}$, $\lambda=1$ only $R_{\mathrm{macro}}$; $\lambda=0.7$
is our default. \textbf{Bold} marks the best value in each column.}
\label{tab:lambda_sweep}
\small
\setlength{\tabcolsep}{4pt}
\renewcommand{\arraystretch}{1.05}
\begin{tabular*}{\textwidth}{@{\extracolsep{\fill}}ccccccccccc@{}}
\toprule
& \multicolumn{2}{c}{MAMO-C}
& \multicolumn{2}{c}{IndustryOR}
& \multicolumn{2}{c}{OptMath}
& \multicolumn{2}{c}{MIPLIB-NL}
& \multicolumn{2}{c}{Average} \\
\cmidrule(lr){2-3}\cmidrule(lr){4-5}\cmidrule(lr){6-7}
\cmidrule(lr){8-9}\cmidrule(lr){10-11}
$\lambda$ & P@1 & P@8 & P@1 & P@8 & P@1 & P@8 & P@1 & P@8 & P@1 & P@8 \\
\midrule
0.0 & \textbf{75.9} & 88.2 & 52.0 & 72.0 & 34.9 & 50.0 & 14.6 & 25.9 & 44.4 & 59.0 \\
0.1 & 72.9 & 88.2 & 52.0 & 72.0 & 36.8 & 52.4 & 15.9 & 32.3 & 44.4 & 61.2 \\
0.3 & 74.9 & 88.2 & 53.0 & 69.0 & 34.3 & 53.0 & 14.6 & \textbf{33.2} & 44.2 & 60.9 \\
0.5 & 74.4 & 88.2 & 50.0 & 70.0 & 33.1 & 53.4 & 14.6 & 32.7 & 43.0 & 61.1 \\
\textbf{0.7} & \textbf{75.9} & \textbf{90.2} & 53.0 & 71.0 & 36.7 & \textbf{54.8} & \textbf{16.8} & \textbf{33.2} & \textbf{45.6} & \textbf{62.3} \\
0.9 & 68.5 & 88.2 & \textbf{55.0} & \textbf{73.0} & 36.6 & 54.4 & 14.1 & 32.3 & 43.6 & 62.0 \\
1.0 & 73.9 & 88.7 & 50.0 & 72.0 & \textbf{37.4} & 53.6 & 14.6 & 31.8 & 44.0 & 61.5 \\
\bottomrule
\end{tabular*}
\end{table}

The default $\lambda=0.7$ gives the best average Pass@1 and Pass@8.
Average Pass@8 is largely insensitive to $\lambda$ once the macro term is
present (60.9--62.3 for $\lambda\geq 0.1$) and drops to 59.0 only when it
is removed ($\lambda=0$), with the loss concentrated on MIPLIB-NL. This
indicates that cross-strategy exploration is what drives repeated-sampling
coverage. Average Pass@1 varies within about 2.6 points across the sweep,
and the per-benchmark fluctuations do not follow a consistent trend, so we
keep $\lambda=0.7$ as the default.

\subsection{Disentangling the Effects of Strategy Routing and Reinforcement Learning}
\label{app:routing_ablation}

To disentangle the effect of reinforcement learning from the effect of
strategy-diverse routing, we compare four configurations of the same
Qwen3-4B-Instruct-2507 backbone under two prompting protocols: the rigid
\textbf{SIR Prompt}, which follows the solver-integrated reasoning
paradigm of prior work~\citep{chen2026solver} and requires every instance
to be formulated as a solver-executable mathematical program, and our
\textbf{Meta Prompt}, which lets the model choose among SIR, Exact
Combinatorial Algorithm, and Heuristic Search according to the problem
structure. \textbf{Base (SIR)} and \textbf{Base (Meta)} evaluate the
untrained backbone under the two prompts. \textbf{SIR-RL} is trained with
the same GRPO recipe and the same text-only data as SDRL but is confined
to the SIR Prompt during both training and inference, without the
diversity reward $R_{\mathrm{div}}$; it isolates the gain from
executable-feedback RL within a single paradigm. \textbf{SDRL} is trained
under the Meta Prompt with the diversity reward on the same text-only
data, and thus corresponds to the Text-only variant in
Table~\ref{tab:mixed_format_ablation} rather than the stronger
mixed-format model in Table~\ref{tab:pass1_accuracy}.
Table~\ref{tab:pdrl_sir_ablation} reports Pass@1 and Pass@8 across all
seven benchmarks.

\begin{table*}[htbp]
\centering
\caption{
Pass@1 and Pass@8 accuracy (\%) for disentangling the effects of strategy routing and reinforcement learning.
\textbf{Base (SIR)} and \textbf{Base (Meta)} evaluate Qwen3-4B-Instruct-2507 under the rigid SIR Prompt and the Meta Prompt, respectively.
\textbf{SIR-RL} applies reinforcement learning while remaining restricted to the SIR paradigm, whereas \textbf{SDRL (Ours)} enables strategy-diverse routing under the Meta Prompt.
}
\label{tab:pdrl_sir_ablation}

\setlength{\tabcolsep}{3.5pt}
\renewcommand{\arraystretch}{1.08}

\resizebox{\textwidth}{!}{%
\scriptsize
\begin{tabular}{lcccccccccccccccc}
\toprule
& \multicolumn{2}{c}{NL4Opt}
& \multicolumn{2}{c}{MAMO-E}
& \multicolumn{2}{c}{MAMO-C}
& \multicolumn{2}{c}{IndustryOR}
& \multicolumn{2}{c}{OptMath}
& \multicolumn{2}{c}{OptiBench}
& \multicolumn{2}{c}{MIPLIB-NL}
& \multicolumn{2}{c}{\textbf{Avg.}} \\

\cmidrule(lr){2-3}
\cmidrule(lr){4-5}
\cmidrule(lr){6-7}
\cmidrule(lr){8-9}
\cmidrule(lr){10-11}
\cmidrule(lr){12-13}
\cmidrule(lr){14-15}
\cmidrule(lr){16-17}

\textbf{Method}
& P@1 & P@8
& P@1 & P@8
& P@1 & P@8
& P@1 & P@8
& P@1 & P@8
& P@1 & P@8
& P@1 & P@8
& P@1 & P@8 \\

\midrule

Base (SIR)
& 91.8 & 95.5
& 89.1 & 94.9
& 41.9 & 66.0
& 48.0 & 63.0
& 12.7 & 32.5
& 62.5 & 70.1
& 8.6 & 25.8
& 50.7 & 64.0 \\

Base (Meta)
& 72.7 & 93.5
& 68.9 & 89.6
& 42.4 & 76.9
& 40.0 & 62.0
& 17.5 & 36.8
& 54.1 & 70.1
& 7.7 & 19.1
& 43.3 & 64.0 \\

\midrule

SIR-RL
& \textbf{94.7} & \textbf{95.9}
& \textbf{92.8} & 96.6
& 70.4 & 85.7
& 48.0 & 65.0
& 36.1 & 48.2
& \textbf{66.8} & 69.6
& 14.1 & 28.2
& 60.4 & 69.9 \\

\textbf{SDRL (Ours)}
& 93.9 & 95.5
& 92.2 & \textbf{97.4}
& \textbf{75.9} & \textbf{90.2}
& \textbf{53.0} & \textbf{71.0}
& \textbf{36.7} & \textbf{54.8}
& 65.8 & \textbf{72.6}
& \textbf{16.8} & \textbf{33.2}
& \textbf{62.0} & \textbf{73.5} \\

\bottomrule
\end{tabular}%
}
\end{table*}

Two observations stand out. First, reinforcement learning is necessary
but not sufficient. SIR-RL improves over Base (SIR) by 9.7 points on
average Pass@1, and SDRL improves over Base (Meta) by 18.7 points, nearly
twice the gain. The base model itself cannot exploit the Meta Prompt:
Base (Meta) trails Base (SIR) by 7.4 points at Pass@1 even though the two
tie at Pass@8. Offering three strategies to an untrained model only adds
ways to go wrong on a single attempt; the routing has to be learned, and
SDRL learns it without any strategy labels.

Second, SDRL solves a strictly harder learning problem than SIR-RL, yet
comes out ahead at both budgets. SIR-RL only has to sharpen one paradigm;
SDRL has to learn when each of three paradigms applies and how to execute
each of them. Despite this, SDRL leads SIR-RL on average Pass@1 (62.0
vs.\ 60.4), winning four of the seven benchmarks while trailing by at
most one point on the remaining three. The gap widens under repeated sampling:
SDRL achieves the higher Pass@8 on six of seven benchmarks and leads by
3.6 points on average, with the largest margins on OptMath (+6.6),
IndustryOR (+6.0), MIPLIB-NL (+5.0), and MAMO-ComplexLP (+4.5). These are
the benchmarks where problem structure varies most across instances, so a
policy that can switch paradigms has the most to gain. Strategy-diverse
training does not merely reshuffle which trajectory ranks first; it
enlarges the set of trajectories that can succeed, and that enlargement is
what repeated sampling exploits.

\begin{figure*}[htbp]
    \centering
    \includegraphics[width=\textwidth]{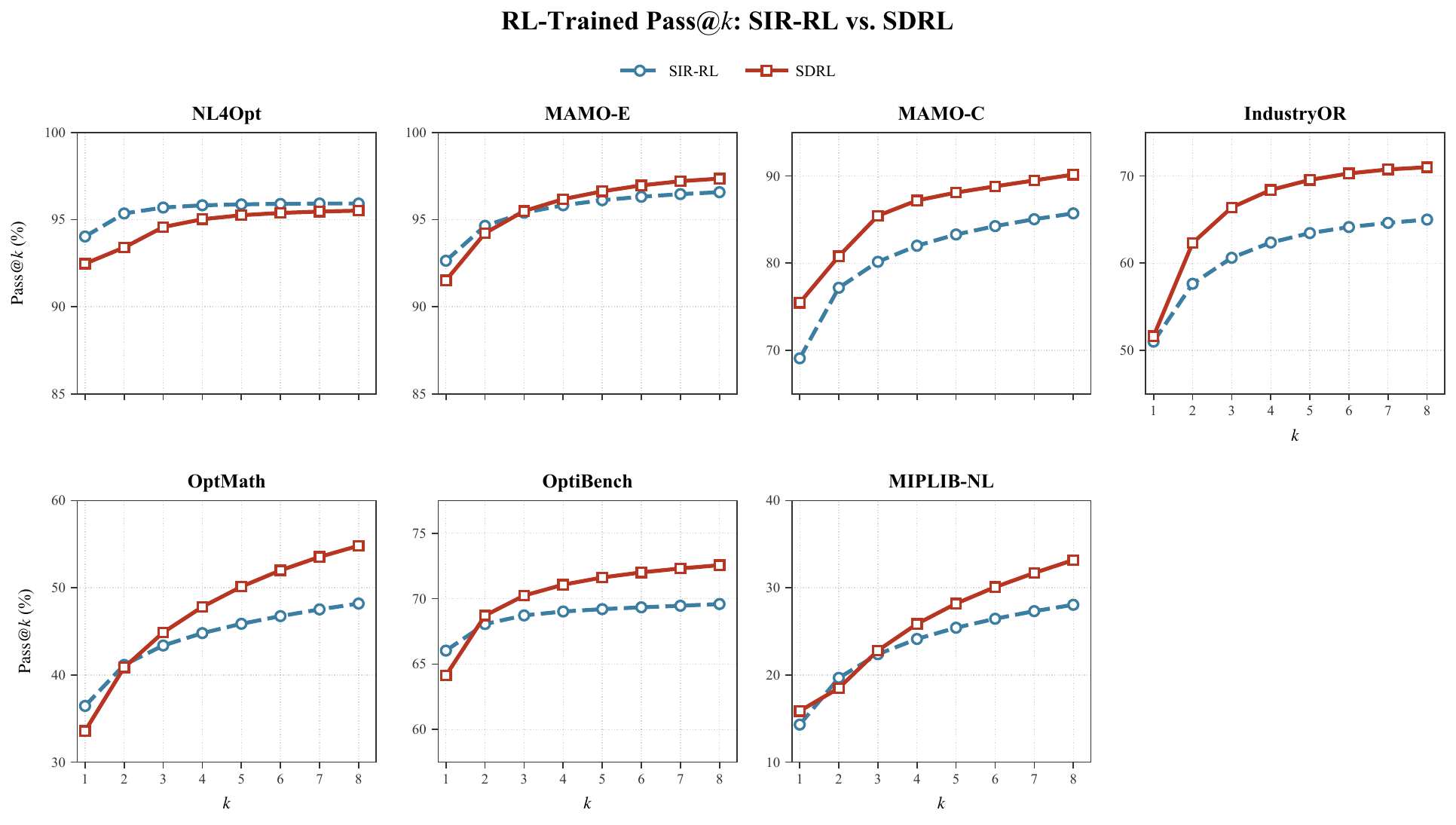}
    \caption{Pass@$k$ of \textbf{SIR-RL} and \textbf{SDRL} across seven
    benchmarks. SDRL's advantage widens with $k$: complementary solving
    pathways turn additional samples into additional solved instances,
    whereas a single-paradigm policy saturates. All points are computed
    from the eight temperature-1.0 samples used for Pass@8.}
    \label{fig:passk_comparison}
\end{figure*}

To trace this effect over the full sampling budget, we plot Pass@$k$ for
$k=1,\ldots,8$ in Figure~\ref{fig:passk_comparison}. The curves are
computed from the same eight temperature-1.0 samples used for Pass@8
(Appendix~\ref{app:eval_settings}), so the $k=1$ point reflects a single
high-temperature sample and is not directly comparable to the Pass@1
values in Table~\ref{tab:pdrl_sir_ablation}, which use temperature 0.5.

The Pass@$k$ curves make the mechanism visible. SIR-RL's curves flatten
early: every additional sample is drawn from the same paradigm, so it
tends to fail on the same instances for the same reasons. SDRL's curves
keep rising, because a sample that switches strategy can succeed where
the previous ones failed. On the benchmarks where SIR-RL leads at small
$k$, SDRL closes the gap and overtakes it as $k$ grows; the only exception
is NL4Opt, where both methods saturate above 95\% and the two curves are
indistinguishable.

Taken together, these results attribute SDRL's gains to the right source.
Reinforcement learning raises trajectory quality within any paradigm;
strategy flexibility alone raises nothing. What SDRL adds is the ability
to route across paradigms and to keep the rollouts diverse enough that
this routing pays off under sampling. The result is a policy that already
outperforms a specialized SIR policy on a single attempt and is markedly
more reliable given several, and the mixed-format model in the main
results extends this advantage further.

\subsection{Runtime Analysis}
\label{app:runtime}

\begin{figure}[h]
    \centering
    \includegraphics[width=0.85\textwidth]{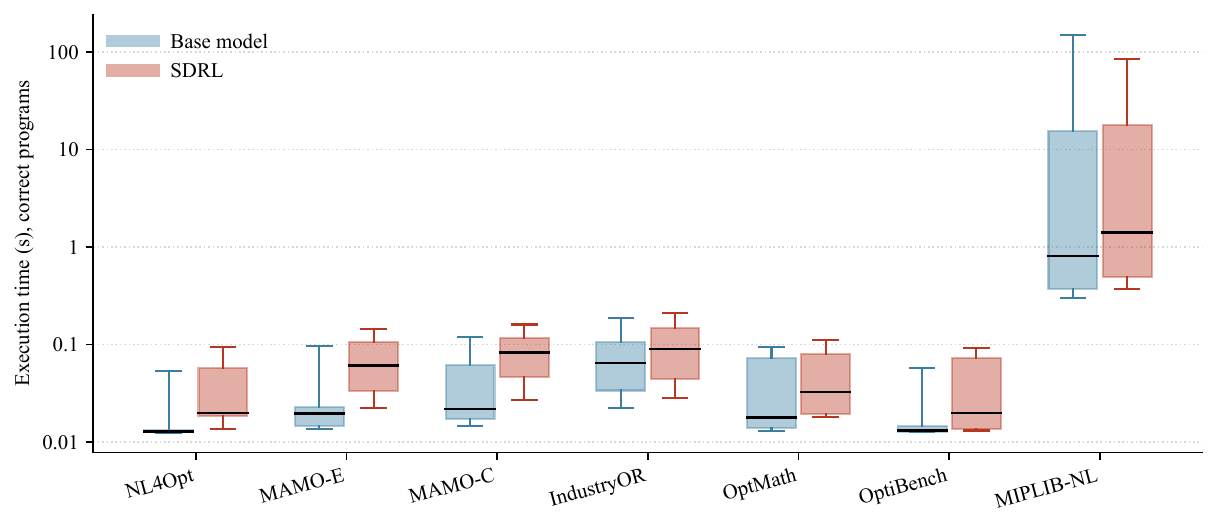}
    \caption{Execution time of correct programs on the Pass@8 rollouts, by
    benchmark (log scale; boxes show quartiles, whiskers the 10th and 90th
    percentiles).}
    \label{fig:runtime_by_benchmark}
\end{figure}

The micro-level diversity reward (Section~\ref{subsec:div_reward_formulation})
rewards rare runtime modes among correct programs, which raises the
question of whether the policy could exploit it by generating artificially
slow programs. We check this by re-executing every Pass@8 rollout of the
Qwen3-4B-Instruct-2507 base model and SDRL-Qwen3-4B in an isolated
subprocess on a dual-socket AMD EPYC 9354 server (64 cores, 128 threads)
with 32 concurrent workers, each limited to a single Gurobi thread, 8\,GB
of memory, and a 1{,}800-second time limit. Runtime is the end-to-end wall
time of the subprocess, including interpreter start-up and imports.
Table~\ref{tab:runtime} and Figure~\ref{fig:runtime_by_benchmark} report
the runtime of programs whose objective value is correct under the
$10^{-6}$ tolerance.

The runtime profile of SDRL is essentially unchanged from the base model.
On the six textual benchmarks both models solve almost every instance in
well under a second, with medians between 0.01 and 0.09\,s and 90th
percentiles below 0.22\,s; the small upward shift in SDRL's medians is
within the fixed interpreter and import overhead. On MIPLIB-NL, the only
benchmark with substantive runtimes, SDRL's 90th percentile is lower than
the base model's (84.5\,s vs.\ 150.4\,s) while it solves 3.4 times as many
rollouts correctly. Timeouts are rare for both models (0.5\% of SDRL
rollouts vs.\ 0.8\% for the base model). The micro-level reward therefore
diversifies runtime modes without inducing slow programs.

\begin{table}[h]
\centering
\small
\caption{Runtime (s) of correct programs on the Pass@8 rollouts. $n$ is the
number of correct programs; Median and P90 are the median and 90th
percentile of their wall-clock execution time.}
\label{tab:runtime}
\begin{tabular}{lrrrrrr}
\toprule
& \multicolumn{3}{c}{\textbf{Base model}} & \multicolumn{3}{c}{\textbf{SDRL}} \\
\cmidrule(lr){2-4}\cmidrule(lr){5-7}
\textbf{Benchmark} & $n$ & Median & P90 & $n$ & Median & P90 \\
\midrule
NL4Opt     & 1443 & 0.013 & 0.05  & 1743 & 0.020 & 0.10 \\
MAMO-E     & 3494 & 0.020 & 0.10  & 4627 & 0.061 & 0.14 \\
MAMO-C     &  689 & 0.022 & 0.12  & 1270 & 0.083 & 0.16 \\
IndustryOR &  314 & 0.065 & 0.19  &  445 & 0.090 & 0.21 \\
OptMath    &  212 & 0.018 & 0.10  &  498 & 0.032 & 0.11 \\
OptiBench  & 3107 & 0.013 & 0.06  & 3254 & 0.020 & 0.09 \\
MIPLIB-NL  &  106 & 0.806 & 150.4 &  360 & 1.409 & 84.5 \\
\bottomrule
\end{tabular}
\end{table}

\section{Case Studies}
\label{app:case_study}

This section presents two qualitative examples illustrating complementary
capabilities of SDRL. The first case demonstrates file-grounded problem solving,
where the model must access external structured data at runtime to construct and
solve the optimization problem. The second case illustrates strategy diversity
on a self-contained textual instance, where SDRL produces correct solutions
through SIR, Exact Combinatorial Algorithm, and
Heuristic Search.

\subsection{File-Grounded Problem Solving}
\label{app:case_file}

We first present a file-grounded instance from MIPLIB-NL to illustrate
SDRL's ability to solve optimization problems whose instance-specific
numerical data are stored in external files. Unlike self-contained textual
instances, the model must access and interpret the associated files at
runtime before constructing the optimization problem.

\begin{tcolorbox}[
    enhanced, breakable,
    colback=gray!8!white, colframe=black,
    coltitle=white, colbacktitle=black,
    title=Problem Statement,
    fonttitle=\bfseries, boxrule=0.8pt, arc=3pt,
    left=6pt, right=6pt, top=3pt, bottom=3pt,
    toptitle=2pt, bottomtitle=2pt, width=\textwidth,
    before skip=4pt, after skip=4pt
]
\begingroup
\scriptsize
\setlength{\parindent}{0pt}
\setlength{\parskip}{1pt}
\raggedright

\textbf{Instance:} \texttt{beasleyC1}

A logistics operator is redesigning a regional backbone with 500 sites and
1,250 candidate directed links. Activating a link incurs a fixed operating
cost, and only activated links can carry flow. Each activated link has a
capacity of 4 units. The objective is to select a subset of links and route
the required flow such that all node-level supply and demand requirements
are satisfied while minimizing the total fixed activation cost.

The instance-specific numerical data are stored in four external CSV files:

\begin{itemize}[leftmargin=*,itemsep=0pt,topsep=2pt]
    \item \texttt{./data/parameter.csv}: global parameters;
    \item \texttt{./data/arcs.csv}: 1,250 candidate directed arcs;
    \item \texttt{./data/costs.csv}: fixed activation costs for the arcs;
    \item \texttt{./data/flow\_balance.csv}: supply/demand requirements for 500 nodes.
\end{itemize}

The generated program must load these files at runtime and determine the
optimal objective value without hard-coding the instance data.

\endgroup
\end{tcolorbox}

SDRL selects the \textbf{SIR} strategy and identifies the problem as
a fixed-charge network flow MILP. It introduces a continuous flow variable
$x_a$ and a binary activation variable $y_a$ for each candidate arc $a$,
minimizes the total activation cost, and enforces node-balance and
arc-capacity constraints.

\begin{tcolorbox}[
    enhanced, breakable,
    colback=optimindred!6!white, colframe=optimindred,
    coltitle=white, colbacktitle=optimindred,
    title=SIR,
    fonttitle=\bfseries, boxrule=0.8pt, arc=3pt,
    left=6pt, right=6pt, top=3pt, bottom=3pt,
    toptitle=2pt, bottomtitle=2pt, width=\textwidth,
    before skip=4pt, after skip=4pt
]
\begingroup
\scriptsize
\setlength{\parindent}{0pt}
\raggedright

\textbf{Strategy tag:}
\texttt{<strategy>SIR</strategy>}

\textbf{Reasoning.}

The model recognizes the instance as a fixed-charge network flow problem.
For each arc, it uses a binary variable to indicate whether the arc is
activated and a continuous variable for the transported flow. The fixed
activation costs form the objective, while flow conservation is enforced
at each node and flow is allowed only on activated arcs.

\textbf{Code (excerpt).}

\begin{Verbatim}[fontsize=\tiny,breaklines=true]
parameter_df = pd.read_csv('./data/parameter.csv')
arcs_df = pd.read_csv('./data/arcs.csv')
costs_df = pd.read_csv('./data/costs.csv')
flow_balance_df = pd.read_csv('./data/flow_balance.csv')

capacity_per_arc = int(
    parameter_df.loc[
        parameter_df['parameter'] == 'capacity_per_arc', 'value'
    ].values[0]
)

for arc_id in arc_ids:
    y_vars[arc_id] = model.addVar(vtype=GRB.BINARY)
    x_vars[arc_id] = model.addVar(lb=0, vtype=GRB.CONTINUOUS)

model.setObjective(
    gp.quicksum(
        cost_data[arc_id]['fixed_cost'] * y_vars[arc_id]
        for arc_id in arc_ids
    ),
    GRB.MINIMIZE
)

for node in flow_balance:
    inflow = gp.quicksum(
        x_vars[a] for a in arc_ids
        if arc_data[a]['end'] == node
    )
    outflow = gp.quicksum(
        x_vars[a] for a in arc_ids
        if arc_data[a]['start'] == node
    )
    model.addConstr(
        inflow - outflow == flow_balance[node]
    )

for arc_id in arc_ids:
    model.addConstr(
        x_vars[arc_id]
        <= capacity_per_arc * y_vars[arc_id]
    )

model.optimize()
objective_value = model.objVal
\end{Verbatim}

\textbf{Result:} objective $=85.0$

\endgroup
\end{tcolorbox}

The generated program successfully loads and integrates all four external
files, constructs the corresponding MILP, and obtains the reference optimum
of \textbf{85}. This example illustrates SDRL's ability to perform
file-grounded optimization without requiring the instance-specific numerical
data to be serialized into the textual prompt.

\newtcolorbox{sirbox}[1][]{
    enhanced, breakable,
    colback=optimindred!6!white, colframe=optimindred,
    coltitle=white, colbacktitle=optimindred,
    title=SIR,
    fonttitle=\bfseries, boxrule=0.8pt, arc=3pt,
    left=6pt, right=6pt, top=3pt, bottom=3pt,
    toptitle=2pt, bottomtitle=2pt, width=\textwidth,
    before skip=4pt, after skip=4pt, #1
}

\newtcolorbox{algobox}[1][]{
    enhanced, breakable,
    colback=algogreen!6!white, colframe=algogreen,
    coltitle=white, colbacktitle=algogreen,
    title=Exact Combinatorial Algorithm,
    fonttitle=\bfseries, boxrule=0.8pt, arc=3pt,
    left=6pt, right=6pt, top=3pt, bottom=3pt,
    toptitle=2pt, bottomtitle=2pt, width=\textwidth,
    before skip=4pt, after skip=4pt, #1
}

\newtcolorbox{heuribox}[1][]{
    enhanced, breakable,
    colback=heurblue!6!white, colframe=heurblue,
    coltitle=white, colbacktitle=heurblue,
    title=Heuristic Search,
    fonttitle=\bfseries, boxrule=0.8pt, arc=3pt,
    left=6pt, right=6pt, top=3pt, bottom=3pt,
    toptitle=2pt, bottomtitle=2pt, width=\textwidth,
    before skip=4pt, after skip=4pt, #1
}

\subsection{Strategy Diversity on a Textual Instance}
\label{app:case_text}

We consider a five-node symmetric traveling salesperson instance: a museum
curator must visit five exhibits exactly once and return to the starting
exhibit, minimizing the total walking distance.

The three responses use distinct computational procedures. The \textbf{SIR}
response formulates an MILP with Miller--Tucker--Zemlin (MTZ) subtour
elimination constraints and solves it with Gurobi. The \textbf{Exact
Combinatorial Algorithm} response fixes the starting exhibit and
exhaustively enumerates the $4!=24$ directed tours. The \textbf{Heuristic
Search} response runs simulated annealing with a 2-opt neighborhood. All
three obtain the optimal tour length of 310.6.

\begin{table}[htbp]
\caption{Three complementary solution strategies for the same five-exhibit
traveling salesperson instance.}
\centering
\small
\begin{tabularx}{\textwidth}{@{}l X c@{}}
\hline
\textbf{Strategy} & \textbf{Representative procedure} & \textbf{Objective} \\
\hline
SIR
& MILP with MTZ subtour elimination, solved with Gurobi & 310.6 \\

Exact Combinatorial Algorithm
& Exhaustive enumeration of the $4!$ tours from a fixed start & 310.6 \\

Heuristic Search
& Simulated annealing with 2-opt moves & 310.6 \\
\hline
\end{tabularx}

\label{tab:case_museum_tsp}
\end{table}

The problem statement and the three responses below illustrate the
reasoning and code produced under each strategy tag; markdown emphasis
in the original outputs has been retyped in \LaTeX{} for readability.

\begin{tcolorbox}[
    enhanced, breakable,
    colback=gray!8!white, colframe=black,
    coltitle=white, colbacktitle=black,
    title=Problem Statement,
    fonttitle=\bfseries, boxrule=0.8pt, arc=3pt,
    left=6pt, right=6pt, top=3pt, bottom=3pt,
    toptitle=2pt, bottomtitle=2pt, width=\textwidth,
    before skip=4pt, after skip=4pt
]
\begingroup
\scriptsize
\setlength{\parindent}{0pt}
\setlength{\parskip}{1pt}
\raggedright
A museum curator is tasked with arranging a tour that visits five distinct
historical exhibits in a sprawling complex. The walking distances between
exhibits are recorded as follows:
\begin{itemize}[leftmargin=*,itemsep=0pt,topsep=2pt]
\item Exhibit Alpha to Beta: 10.0 meters
\item Exhibit Alpha to Gamma: 129.8 meters
\item Exhibit Alpha to Delta: 40.3 meters
\item Exhibit Alpha to Epsilon: 14.0 meters
\item Exhibit Beta to Gamma: 137.3 meters
\item Exhibit Beta to Delta: 50.0 meters
\item Exhibit Beta to Epsilon: 18.0 meters
\item Exhibit Gamma to Delta: 115.7 meters
\item Exhibit Gamma to Epsilon: 138.8 meters
\item Exhibit Delta to Epsilon: 37.1 meters
\end{itemize}
The objective is to determine the minimal total walking path that allows
the curator to stop at each exhibit precisely once and then return to the
initial exhibit.
\endgroup
\end{tcolorbox}

\begin{sirbox}
\begingroup\scriptsize\setlength{\parindent}{0pt}\raggedright
\textbf{Strategy:} SIR

\textbf{Raw strategy tag:}
\texttt{<strategy>SIR</strategy>}

\textbf{Reasoning.}

\textbf{Mathematical Analysis and Formulation.}
This problem is a classic \textbf{Traveling Salesperson Problem (TSP)},
where the objective is to find the shortest possible route that visits
every exhibit exactly once and returns to the origin.

\textit{\textbf{1. Sets and Indices:}}
\begin{itemize}[leftmargin=*,itemsep=0pt,topsep=2pt]
\item Let $V=\{0,1,2,3,4\}$ be the set of nodes representing the exhibits
Alpha, Beta, Gamma, Delta, and Epsilon respectively.
\item Node $0$ is designated as the starting/ending point (Alpha) for
subtour elimination purposes.
\end{itemize}

\textit{\textbf{2. Parameters:}}
$d_{ij}$: the walking distance between exhibit $i$ and exhibit $j$. The
distance matrix is symmetric, $d_{ij}=d_{ji}$.

\textit{\textbf{3. Decision Variables:}}
\begin{itemize}[leftmargin=*,itemsep=0pt,topsep=2pt]
\item $x_{ij}\in\{0,1\}$ for all $i,j\in V$, $i\neq j$: equals 1 if the
path proceeds directly from exhibit $i$ to exhibit $j$, and 0 otherwise.
\item $u_i\geq 1$ for all $i\in V\setminus\{0\}$: continuous variables
used for the Miller--Tucker--Zemlin (MTZ) subtour elimination constraints,
representing the sequence in which the exhibits are visited.
\end{itemize}

\textit{\textbf{4. Objective Function:}}
Minimize the total walking distance of the tour,
$\min \sum_{i\in V}\sum_{j\in V,\,j\neq i} d_{ij}x_{ij}$.

\textit{\textbf{5. Constraints:}}
\begin{itemize}[leftmargin=*,itemsep=0pt,topsep=2pt]
\item \textbf{Leave exactly once (out-degree):}
$\sum_{j\neq i} x_{ij}=1$ for all $i\in V$.
\item \textbf{Arrive exactly once (in-degree):}
$\sum_{i\neq j} x_{ij}=1$ for all $j\in V$.
\item \textbf{Subtour elimination (MTZ):}
$u_i-u_j+|V|\,x_{ij}\leq |V|-1$ for all $i,j\in V\setminus\{0\}$,
$i\neq j$, ensuring a single continuous tour with no disconnected loops
among nodes $\{1,2,3,4\}$.
\end{itemize}

\textit{\textbf{6. Size Estimation:}}
$5\times 4=20$ binary $x_{ij}$ variables and $4$ continuous $u_i$
variables, i.e., 24 decision variables; 10 degree constraints plus
$4\times 3=12$ MTZ constraints, i.e., 22 constraints in total.

\textbf{Code (excerpt).}
\begin{Verbatim}[fontsize=\tiny,breaklines=true]
model = gp.Model("Museum_TSP", env=env)
x = {(i, j): model.addVar(vtype=GRB.BINARY) for i in nodes for j in nodes if i != j}
u = {i: model.addVar(lb=1, ub=n-1, vtype=GRB.CONTINUOUS) for i in nodes[1:]}
model.setObjective(gp.quicksum(d[i, j]*x[i, j] for (i, j) in x), GRB.MINIMIZE)
for i in nodes:
    model.addConstr(gp.quicksum(x[i, j] for j in nodes if i != j) == 1)
    model.addConstr(gp.quicksum(x[j, i] for j in nodes if i != j) == 1)
for i in nodes[1:]:
    for j in nodes[1:]:
        if i != j:
            model.addConstr(u[i] - u[j] + n*x[i, j] <= n - 1)
model.optimize()
objective_value = model.ObjVal
\end{Verbatim}

\textbf{Result:} objective $=$ 310.6
\endgroup
\end{sirbox}

\begin{algobox}
\begingroup\scriptsize\setlength{\parindent}{0pt}\raggedright
\textbf{Strategy:} Exact Combinatorial Algorithm

\textbf{Raw strategy tag:}
\texttt{<strategy>Exact Combinatorial Algorithm</strategy>}

\textbf{Reasoning.}

\textit{\textbf{1. Core Algorithmic Structure \& Problem Properties.}}
The problem asks for the shortest path visiting five distinct exhibits
exactly once and returning to the starting point. This is the classic
symmetric Traveling Salesperson Problem (TSP). The exhibits are the nodes
in a fully connected undirected graph, and the walking distances are the
edge weights. The goal is to find a Hamiltonian cycle of minimum total
weight. While TSP is NP-hard in general---meaning its exact solution
typically requires dynamic programming (like the Held--Karp algorithm with
$O(n^2 2^n)$ time complexity) or branch-and-bound algorithms---the problem
size here is exceedingly small ($N=5$).

\textit{\textbf{2. Algorithm Selection: Exhaustive Search.}}
Because $N=5$, the number of possible tours is $(N-1)!/2=12$ unique
undirected cycles for a symmetric graph. Even generating all $4!=24$
directed cycles by fixing a starting node is computationally trivial.
Therefore, exhaustive search is the most robust and provably correct method
for this scale: it guarantees global optimality without the overhead of
search-tree pruning or memoization structures.

\textit{\textbf{3. Complexity Analysis.}}
\begin{itemize}[leftmargin=*,itemsep=0pt,topsep=2pt]
\item \textbf{Time:} $O((N-1)!)$; for $N=5$ this is 24 path evaluations,
each requiring $O(N)$ operations to sum the edge distances.
\item \textbf{Space:} $O(N^2)$ for the distance matrix and $O(N)$ for the
current permutation and best sequence.
\end{itemize}

\textit{\textbf{4. Variable Representation.}}
The tours are evaluated implicitly. Once the optimal sequence is
identified, it is mapped to binary decision variables $x_{i,j}$, where
$x_{i,j}=1$ if the curator walks directly from exhibit $i$ to exhibit $j$,
and $0$ otherwise.

\textbf{Code (excerpt).}
\begin{Verbatim}[fontsize=\tiny,breaklines=true]
start_node = "Alpha"
other_nodes = [v for v in nodes if v != start_node]
best_cost, best_tour = float('inf'), None
for perm in itertools.permutations(other_nodes):
    tour = [start_node] + list(perm) + [start_node]
    cost = sum(dist_matrix[tour[i]][tour[i+1]] for i in range(len(tour) - 1))
    if cost < best_cost:
        best_cost, best_tour = cost, tour
objective_value = best_cost
\end{Verbatim}

\textbf{Result:} objective $=$ 310.6
\endgroup
\end{algobox}

\begin{heuribox}
\begingroup\scriptsize\setlength{\parindent}{0pt}\raggedright
\textbf{Strategy:} Heuristic Search

\textbf{Raw strategy tag:}
\texttt{<strategy>Heuristic Search</strategy>}

\textbf{Reasoning.}

\textbf{Heuristic Design: Simulated Annealing (SA).}

\textit{\textbf{1. Optimization Structure and Difficulty.}}
The problem is a symmetric TSP on a complete graph with $n=5$ nodes. The
objective is to find a Hamiltonian cycle of minimum weight. While $n=5$ is
a trivially small instance, the general TSP is NP-hard due to the factorial
growth of permutations. The main source of combinatorial difficulty is
sequencing the exhibits so as to minimize the total wrap-around distance;
feasibility requires visiting every node exactly once and returning to the
origin.

\textit{\textbf{2. Chosen Method.}}
We implement \textbf{Simulated Annealing (SA)}. SA starts from an initial
solution and iteratively explores the solution space; by accepting
non-improving solutions with a probability that decreases over time (the
``temperature''), it escapes local optima. SA is not exact and does not
prove global optimality, but it almost always converges to the optimal or a
highly competitive near-optimal solution for small to medium routing
problems.

\textit{\textbf{3. Solution Representation and Initialization.}}
A permutation of the five exhibit names, with the return step implicit in
the cost; the initial tour is a random shuffle.

\textit{\textbf{4. Neighborhood Operator.}}
\textbf{2-opt:} two distinct indices are selected at random and the segment
between them is reversed. Every neighbor remains a valid Hamiltonian
cycle, so no penalty functions or repair steps are required.

\textit{\textbf{5. Acceptance Rule (Metropolis).}}
With $\Delta E=\text{new distance}-\text{current distance}$, a move is
accepted automatically if $\Delta E<0$, and with probability
$\exp(-\Delta E/T)$ otherwise.

\textit{\textbf{6. Stopping Criterion and Random Seed.}}
Geometric cooling $T_{k+1}=\alpha T_k$ with $T_0=100$, $\alpha=0.99$, and
20 moves per temperature, stopping when $T<10^{-4}$; the random seed is
fixed to 42 for reproducibility.

\textbf{Code (excerpt).}
\begin{Verbatim}[fontsize=\tiny,breaklines=true]
T, T_min, alpha, iter_per_temp = 100.0, 1e-4, 0.99, 20
while T > T_min:
    for _ in range(iter_per_temp):
        i, j = sorted(random.sample(range(len(nodes)), 2))
        new_tour = current_tour[:]
        new_tour[i:j+1] = reversed(new_tour[i:j+1])      # 2-opt move
        delta = calc_tour_distance(new_tour) - current_cost
        if delta < 0 or random.random() < math.exp(-delta / T):
            current_tour, current_cost = new_tour, current_cost + delta
            if current_cost < best_cost:
                best_tour, best_cost = current_tour[:], current_cost
    T *= alpha
objective_value = best_cost
\end{Verbatim}

\textbf{Result:} objective $=$ 310.6
\endgroup
\end{heuribox}

\end{document}